\documentclass[final,authoryear,3p,times]{elsarticle}

\usepackage{graphicx,color}
\usepackage{enumerate}
\usepackage[colorlinks]{hyperref}
\usepackage{natbib}
\usepackage{amsmath,amssymb,amsthm,bm}
\usepackage{multicol}
\usepackage[dvipsnames]{xcolor}
\usepackage{ulem}
\usepackage{tikz}
\usetikzlibrary{automata,positioning}
\usepackage{afterpage}
\usepackage{calc}
\usepackage{ifthen}
\usepackage{algorithm}
\usepackage{algpseudocode}
\usepackage{caption}
\usepackage{subcaption}
\usepackage{tabularx,ragged2e}
\usepackage{colortbl}
\usepackage{amssymb}
\usepackage[title]{appendix}
\usepackage{cancel}
\usepackage{ulem}

\journal{journal}
\newcommand{\boldface}[1]{\boldsymbol{#1}}  

\newcommand{\bfu}{\boldface{u}}

\newcommand{\bfw}{\boldface{w}}

\newcommand{\bfz}{\boldface{z}}

\newcommand{\bfC}{\boldface{C}}

\newcommand{\bfF}{\boldface{F}}

\newcommand{\bfI}{\boldface{I}}

\newcommand{\bfP}{\boldface{P}}

\newcommand{\calB}{\mathcal{B}}

\newcommand{\calD}{\mathcal{D}}

\newcommand{\calQ}{\mathcal{Q}}

\newcommand{\calU}{\mathcal{U}}

\newcommand{\calW}{\mathcal{W}}

\newcommand{\Rset}{\mathbb{R}}

\newlength{\boxwidth}
\def\dd{\;\!\mathrm{d}}

\def\btheorem{\begin{theorem}}
\def\etheorem{\end{theorem}}
\def\blemma{\begin{lemma}}
\def\elemma{\end{lemma}}
\def\bproposition{\begin{proposition}}
\def\eproposition{\end{proposition}}
\def\bcorollary{\begin{corollary}}
\def\ecorollary{\end{corollary}}
\def\bdefinition{\begin{definition}}
\def\edefinition{\end{definition}}
\def\bexample{\begin{example}}
\def\eexample{\end{example}}
\def\bremark{\begin{remark}}
\def\eremark{\end{remark}}

\newcommand{\eps}{\varepsilon}

\DeclareMathOperator{\tr}{tr}

\newcommand{\be}{\begin{equation}}
\newcommand{\ee}{\end{equation}}
\newcommand{\beq}{\begin{eqnarray}}
\newcommand{\eeq}{\end{eqnarray}}
\newcommand{\bem}{\begin{multline}}
\newcommand{\eem}{\end{multline}}
\newcommand{\ba}{\begin{align}}
\newcommand{\ea}{\end{align}}

\renewcommand{\figurename}{Figure}
\renewcommand{\tablename}{Table}

\newcommand{\RV}[1]{{#1}}

\begin{document}

\begin{frontmatter}

\title{Can KAN CANs? Input-convex Kolmogorov-Arnold Networks (KANs) as  hyperelastic constitutive artificial neural networks (CANs) }

\author[a]{Prakash Thakolkaran\corref{cor1}}
\author[a]{Yaqi Guo\corref{cor1}}
\author[a]{Shivam Saini}
\author[b]{Mathias Peirlinck}
\author[a,b]{Benjamin Alheit}
\author[a]{Siddhant Kumar\corref{cor2}}

\cortext[cor1]{Equal contributions}
\cortext[cor2]{Email: Sid.Kumar@tudelft.nl}

\address[a]{Department of Materials Science and Engineering, Delft University of Technology, 2628 CD Delft, The Netherlands}
\address[b]{Department of Biomechanical Engineering, Delft University of Technology, 2628 CD Delft, The Netherlands}


\begin{abstract}
Traditional constitutive models rely on hand-crafted parametric forms with limited expressivity and generalizability, while neural network-based models can capture complex material behavior but often lack interpretability. To balance these trade-offs, we present  \RV{monotonic} Input-Convex Kolmogorov-Arnold Networks (ICKANs) for learning polyconvex hyperelastic constitutive laws. ICKANs leverage the Kolmogorov-Arnold representation, decomposing the model into compositions of trainable univariate spline-based activation functions for rich expressivity. \RV{We introduce trainable monotonic input-convex splines within the KAN architecture, ensuring physically admissible polyconvex models for isotropic compressible hyperelasticity.}  The resulting models are both compact and interpretable, enabling explicit extraction of analytical constitutive relationships through a \RV{monotonic} input-convex symbolic regression technique. Through unsupervised training on full-field strain data and limited global force measurements, ICKANs accurately capture nonlinear stress–strain behavior across diverse strain states. Finite element simulations of unseen geometries with trained ICKAN hyperelastic constitutive models confirm the framework’s robustness and generalization capability.

\end{abstract}

\end{frontmatter}

\section{Introduction}

Constitutive modeling of material behavior has traditionally relied on \textit{a priori} hand-crafted parametric models, with model parameters iteratively calibrated using simple experimental data, such as tension or torsion tests. However, this approach poses challenges in limited generalization of the constitutive model beyond the calibration data and relies on an inefficient, time-consuming trial-and-error process of hand-crafting such parametric models. In recent years, machine learning (ML) and data-driven techniques have revolutionized traditional phenomenological \RV{constitutive modeling} approaches. Rather than relying on hand-crafted parametric models, these methods extract patterns directly from experimental or multiscale simulation data, enabling the efficient development of physically-consistent, accurate, and generalizable constitutive models \citep{Fuhg2024,kumar_ml_mech_review}. Among the modern data-driven approaches, several categories exist, each with its own advantages and limitations. Here, we provide a brief and non-exhaustive review of recent representative works (see \cite{Fuhg2024} for detailed review) to identify existing gaps and highlight the motivation for this study.

The \textit{model-free} approaches \citep{kirchdoerfer_data-driven_2016,kirchdoerfer_data-driven_2018,eggersmann_model-free_2019,ibanez_data-driven_2017,conti_data-driven_2018,nguyen_data-driven_2018,carrara_data-driven_2020,karapiperis_data-driven_2021} rely on a large catalog of stress-strain pairs and avoid potential errors arising from modeling assumption. In this method, when a new strain query is made, the closest stress-strain pair from the existing dataset is identified, subject to physical compatibility constraints. However, it struggles with poor generalization to stress-strain states beyond the available dataset, lacks the ability to correct for measurement noise, and requires a large amount of data.

In contrast, \textit{model-based} approaches aim to learn physically admissible and consistent surrogate mappings of stress-strain data using various modeling techniques. These methods strive to enhance generalizability, correct for measurement noise, and achieve accurate predictions with minimal data. However, among model-based approaches, state-of-the-art techniques span a broad spectrum in terms of interpretability (i.e., understanding how and why a model makes a certain prediction), generalizability (i.e., extending beyond training data), and model expressivity (i.e., the ability to approximate diverse data distributions and constitutive behaviors). 

On one end of the spectrum are techniques based on sparse and symbolic regression. \textit{Sparse regression} techniques \citep{flaschel_unsupervised_2021,joshi_Bayesian, flaschel_plasticity,Flaschel2023_gsm,Flaschel2023_brain,Marino2023,Wang2021} assume a large catalog of hand-crafted candidate functions and use deterministic or Bayesian methods to select a subset, creating parsimonious and interpretable constitutive models.  \textit{Symbolic regression} techniques \citep{Bahmani2024a,Bahmani2024b,Phan2025,Hou2024,Abdusalamov2023,Kissas2024} explore permutation and combination of mathematical operations on input data to yield an analytical model that best fits the data.  While these approaches offer high interpretability and generalizability as well as analytically verifiable compliance with physical laws, their expressive power is inherently constrained by the predefined set of candidate functions and mathematical operations.

On the other end of the spectrum are techniques based on \textit{Gaussian process regression} (GPR) and \textit{neural networks} (NNs) which are uninterpretable and black-box in nature, but offer higher expressivity. GPR is particularly well-suited for constitutive modeling \citep{Upadhyay2024, rocha_onthefly_2021, fuhg_local_2022, Aggarwal2023GPplanar} in the low-data regime but encounters computational scaling bottlenecks as the amount of data increases. Neural networks (NNs) as constitutive models overcome this bottleneck by leveraging scalable, highly parameterized architectures (ranging from thousands to millions of trainable parameters) while providing significantly greater expressivity than GPR. NN-based constitutive models have been successfully developed for a wide range of material behaviors -- from hyperelasticity \citep{Linden2023,Vlassis2020}, viscoelasticity \citep{Abdolazizi2024}, and plasticity \citep{Bonatti2021} to multi-physics phenomena \citep{Klein2024,Kalina2024}. These models employ diverse architectures, including recurrent neural networks \citep{Mozaffar2019}, long short-term memory networks, hypernetworks \citep{Zheng2024}, neural ordinary differential equations \citep{Tac2022}, sparse networks with interpretable hidden features \citep{Linka2023}, Bayesian neural networks \citep{Linka2025}, probabilistic diffusion fields \citep{Tac2024}, and many more; see \cite{Fuhg2024} for an exhaustive review.

To strike a balance between interpretability and expressivity -- in the general ML setting and independent of constitutive modeling -- \textit{Kolmogorov-Arnold Networks} (KANs) \citep{liu2024kan} have recently emerged as a novel NN architecture.  The Kolmogorov-Arnold representation theorem \citep{Schmidt-Hieber2020} states that any multivariate continuous function on a bounded domain can be written as a finite composition of continuous univariate functions and the binary operation of addition. KANs take advantage of the Kolmogorov-Arnold representation theorem to break down high-dimensional functions into simpler univariate functions. Like standard multi-layer perceptrons (MLPs), KANs are fully connected networks. However, they differ in a key way. MLPs learn by adjusting weights and biases, with fixed activation functions providing nonlinearity. In contrast, in KANs, the activation functions themselves are  trainable parameters, which allows the network to directly learn the nonlinear transformations. Since their introduction, KANs have gained significant attention and have been extended to solving forward and inverse problems as well as operator learning for partial differential equations \citep{patra2024physics,ABUEIDDA2025117699, WANG2025117518,Gao2025,Toscano2024}.   In this paper, we first address the question of \uline{how KANs can be adapted for constitutive modeling, focusing on hyperelasticity}?

A major challenge in adapting NNs, including KANs, for constitutive modeling is their high expressivity---an advantage that enables them to capture complex material behavior but, if left unchecked, can lead to unphysical predictions of constitutive behavior. To address this, significant efforts have been made to incorporate physical knowledge into NN architectures, constraining them to the subspace of physically admissible constitutive models and mitigating their tendency to overfit or produce non-physical responses. In the context of hyperelasticity, physical admissibility conditions include, for example, (poly)convexity of the strain energy density, objectivity, and zero stress at zero deformation \citep{Linden2023}. Several recent studies have tackled this challenge using various adaptations of the \textit{input-convex neural network} (ICNN) architecture \citep{amos_input_2017} for modeling hyperelastic strain energy densities \citep{thakolkaran2022nn,Asad2022,klein_polyconvex_2022,Meng2025,Shi2025,Jailin2024}.  The ICNN architecture is based on adapting a vanilla MLP architecture to ensure that the output is convex with respect to the input values, which aligns well with the (poly)convexity requirements of a hyperelastic strain energy density. However, the ICNN architecture is not directly applicable to KANs due to fundamental differences in their formulation. 
\RV{We note that an independent study by \citet{Abdolazizi2025} has also proposed using KANs as hyperelastic constitutive NNs. In contrast to our approach, the KAN models presented by \citet{Abdolazizi2025} are not input-convex and thus do not satisfy the (poly-)convexity requirements of hyperelasticity. Additionally, their approach was limited to supervised training using only stress-strain pairs. As discussed below, unsupervised training and physically admissible generalization is not feasible without enforcing (poly-)convexity.}
In this study, we explore \uline{how to build a \RV{monotonic} Input-Convex KAN (ICKAN) architecture, thereby enabling physically admissible KANs-based hyperelastic constitutive models} for balanced interpretability and expressivity. 

Another challenging aspect of hyperelastic constitutive NNs is their training. While training on labeled stress-strain pairs from multiscale simulations is straightforward \RV{(commonly done using Sobolev training; see e.g., \citet{Bahmani2024a})}, most common experimental setups (e.g., uniaxial or biaxial tension and torsion tests) fail to sufficiently probe the high-dimensional stress-strain space required to train highly parameterized NNs. Additionally, full-field experimental tests using digital image correlation (DIC) provide strain fields but only boundary-averaged projections of stress tensors (i.e., reaction forces). Training constitutive NNs on full-field strain fields and reaction forces without explicit stress labels presents an unsupervised learning challenge.\footnote{In a typical DIC setup, although the strain field is discretized at thousands-to-millions of points, only one or two reaction forces can be measured. As a result, reaction forces act as a regularization rather than serving as direct labels for supervised training.}  Here, we address the question of \uline{how to train ICKAN-based hyperelastic constitutive models in an unsupervised manner}.

To tackle this challenge, we previously introduced the  NN-EUCLID framework \citep{thakolkaran2022nn}, demonstrating that NN-based hyperelastic constitutive NNs can be trained without stress data, relying solely on strain fields and global force measurements--quantities readily available through mechanical testing and DIC. Without stress labels, the learning process is guided by a physics-motivated loss function based on the conservation of linear momentum. NN-EUCLID builds upon the earlier EUCLID framework \citep{flaschel_unsupervised_2021,flaschel_plasticity,Flaschel2023_brain,Flaschel2023_gsm,Marino2023,joshi_Bayesian} for unsupervised model discovery via sparse regression, as well as the Virtual Fields Method \citep{Grdiac2006,Pierron2012} for unsupervised calibration of parametric constitutive models. \RV{We note that the Equilibrium Gap Method (EGM) \citep{claire2004finite} and NN-EUCLID are closely related to the Virtual Fields Method (VFM), with EGM being a special case where local ansatz functions are used for the virtual fields. NN-EUCLID employs neural networks to model the constitutive behavior, while using the same ansatz functions for the virtual fields as in EGM, thereby offering a more flexible and modeling-bias-free approach to constitutive modeling.} \RV{The validity of NN-EUCLID} has been further supported by experimental studies conducted by \citet{Jailin2024} and \citet{Meng2025}. Here, we demonstrate that ICKAN-based hyperelastic constitutive models are compatible with NN-EUCLID for unsupervised training, thereby paving the way for their application in realistic experimental settings. \RV{Notably, we demonstrate that  monotonicity and input-convexity in ICKANs are essential for addressing the ill-posedness of unsupervised training---without them, the method fails.}

\section{Modeling hyperelasticity using ICKANs}

\subsection{Hyperelasticity preliminaries}

For hyperelastic materials, the constitutive model is characterized by a strain energy density function $W(\bfF)$, from which both the first Piola-Kirchhoff stress $\bfP(\bfF)$, and the incremental tangent modulus $\mathbb{C}(\bfF)$, are derived:
\be\label{eq:PK-general}
    \bfP(\bfF) = \frac{\partial W(\bfF)}{\partial \bfF}, \quad 
    \mathbb{C}(\bfF) = \frac{\partial \bfP(\bfF)}{\partial \bfF}, 
    \quad \forall \ \bfF \in \text{GL}_+(3).
\ee
Here, $\text{GL}_+(3)$ denotes the set of all invertible second-order tensors with positive determinants. The constitutive modeling task is to choose an appropriate form of $W(\bfF)$ that not only captures the material response accurately but also satisfies key physical and thermodynamic requirements. These requirements include:

\begin{itemize}
    \item \textit{Stress-free undeformed configuration:} In the undeformed state, the stress must vanish:
    \be\label{eq:stress-free}
        \boldsymbol{P}(\boldsymbol{F} = \bfI) = \boldsymbol{0}.
    \ee
    \item \textit{Objectivity:} The strain energy density must be objective, i.e,
    \be
        W(\boldsymbol{R}\boldsymbol{F}) = W(\boldsymbol{F}), \quad \forall \ \boldsymbol{F} \in \text{GL}_+(3), \quad \boldsymbol{R} \in \text{SO}(3),
        \label{objectivity-condition}
    \ee
    where $\text{SO}(3)$ is the group of all 3D rotation matrices. \RV{Note that objectivity implies compatibility with the balance of angular momentum.}
    \item \textit{Polyconvexity:} Material stability is ensured if $W(\bfF)$ satisfies the quasiconvexity condition  \citep{ball_convexity_1976,Schroder2010,Morrey1952}:
    \be
      \RV{\int_\calB W(\bar\bfF + \nabla \bfw)\dd V \geq \int_\calB W(\bar \bfF) \dd V, 
      \quad \forall \ \calB\subset\Rset^3, \ \bar\bfF\in\text{GL}_{+}(3), \ 
      \bfw\in C^\infty_0(\calB),}
    \ee
    with $\bfw$ vanishing on $\partial\calB$. However, direct enforcement of quasiconvexity is generally impractical  \citep{Kumar2019}. Instead, polyconvexity is preferred over quasiconvexity because it offers a more analytically tractable condition and given that polyconvexity inherently implies quasiconvexity. The strain energy density is polyconvex if and only if there exists a convex function \RV{$\mathcal{P}: \mathbb{R}^{3\times3}\times\mathbb{R}^{3\times3}\times \mathbb{R} \rightarrow \mathbb{R}$} such that
    \begin{align}\label{eq:polyconvex}
        W(\boldsymbol{F}) = \mathcal{P}(\boldsymbol{F}, \text{Cof} \ \boldsymbol{F}, \det \boldsymbol{F}).
    \end{align}
\end{itemize}

\subsection{Hyperelastic model ansatz}

For the scope of this study, we consider isotropic compressible hyperelasticity. Incorporating the aforementioned physical constraints, we propose the following ansatz for the strain energy density function:  
\be\label{eq:ansatz}
\begin{aligned}
    W(\bfF) \ &=\  \underbrace{W^{\text{ICKAN}}_{\calQ}\ (K_1,K_2,K_3)}_{\substack{\text{\RV{monotonic} input-convex}\\\text{KAN-based model}}}  \ +  \underbrace{W^\text{0}}_{\substack{\text{Energy}\\\text{correction}}} \ , \\
    & \text{with} \qquad  K_1 = (\Tilde I_1 - 3),\quad K_2 = (I_2^* - 3\sqrt{3}),\quad K_3 = (J-1)^2.
\end{aligned}
\ee  
Here, $W^{\text{ICKAN}}_{\calQ}$ represents a \RV{monotonic} Input-Convex Kolmogorov-Arnold Network (ICKAN) with the trainable parameter set $\calQ$. The architecture of ICKAN is discussed in detail in Section \ref{sec:ICKAN}.  To model isotropic compressible hyperelasticity, we use deviatoric and volumetric invariants, i.e., 
\be 
\begin{aligned} 
\text{deviatoric invariants}: \qquad & \tilde I_1=J^{-2/3}I_1, \quad  I_2^*=(\tilde I_2)^{3/2} = (J^{-4/3}I_2)^{3/2},\\
\text{volumetric invariant}: \qquad & J=\det(\bfF)=I_3^{1/2},
\end{aligned} \ee where
\be I_1=\tr(\bfC),\qquad I_2 = \frac{1}{2}\left[\tr(\bfC)^2-\tr(\bfC^2)\right], \qquad I_3 = \det(\bfC) 
\ee
represent the principal invariants of the right Cauchy-Green deformation tensor $\bfC=\bfF^T\bfF$. $W^0$ is a constant scalar correction such that the energy density vanishes at zero deformation ($\bfF=\bfI$), i.e.,
\be\label{zero-energy}
W(\bfI) = 0 \qquad \implies \qquad W^0 = \ -\  W^{\text{ICKAN}}_{\calQ}(0,0,0).
\ee
Note that the energy correction $W^0$ is updated during each iteration of the training (discussed later in Section \ref{sec:Problem}). \RV{While we consider hyperelastic models with invariants-based inputs here, models based on other inputs such as principal stretches are also possible; see \cite{Abdolazizi2024}.}

The first Piola-Kirchhoff stress and tangent modulus are given by (from \eqref{eq:PK-general}) 
\be\label{eq:PK_ansatz}
P_{ij}(\bfF) = \frac{\partial W^{\text{ICKAN}}_{\calQ}(K_1,K_2,K_3)}{\partial F_{ij}}
\ee
and
\be\label{eq:tangent}
\mathbb{C}_{ijkl} = \frac{\partial P_{ij}(\bfF)}{\partial F_{kl}} = \frac{\partial^2 W_{\calQ}^\text{ICKAN}(K_1,K_2,K_3)}{\partial F_{ij}\partial F_{kl}},
\ee
respectively. Here, we use the Einstein index notation over the subscripts.

We now discuss how this ansatz satisfies the previously mentioned physical constraints.

\begin{itemize}
    \item \textit{Stress-free undeformed configuration}: The inputs $(K_1,K_2,K_3)$ in \eqref{eq:ansatz} are the modified version of the invariants obtained by  appropriately shifting---and in some cases squaring---such that both their values and their derivatives with respect to $\bfF$ vanish in the undeformed state $(\bfF=\bfI)$, i.e.,
    \be
    \frac{\partial K_1}{\partial \bfF}\bigg |_{\bfF=\bfI} = \boldsymbol{0},\qquad \frac{\partial K_2}{\partial \bfF}\bigg |_{\bfF=\bfI} = \boldsymbol{0},\qquad
    \frac{\partial K_3}{\partial \bfF}\bigg |_{\bfF=\bfI} = \boldsymbol{0}.
    \ee
    Consequently, the first Piola-Kirchhoff stress vanishes identically at undeformed state:
    \be\label{zero-stress}
    \bfP(\bfI) = 
    \frac{\partial W^{\text{ICKAN}}_{\calQ}(0,0,0)}{\partial K_1}\frac{\partial K_1}{\partial \bfF}\bigg |_{\bfF=\bfI} 
    +
    \frac{\partial W^{\text{ICKAN}}_{\calQ}(0,0,0)}{\partial K_2}\frac{\partial K_2}{\partial \bfF}\bigg |_{\bfF=\bfI} 
    +
    \frac{\partial W^{\text{ICKAN}}_{\calQ}(0,0,0)}{\partial K_3}\frac{\partial K_3}{\partial \bfF}\bigg |_{\bfF=\bfI}  = \boldsymbol{0}
    \ee
    \item \textit{Objectivity}: Since the model ansatz \eqref{eq:ansatz} is a function of $K_1$, $K_2$, and $K_3$ which are in turn function of invariants of $\bfC$, it is objective by construction.
    \item \textit{Polyconvexity:} The invariant $\tilde I_1$ and equivalently, $K_1$ are polyconvex in $\bfF$. $K_3=(J-1)^2$ is polyconvex in $\bfF$. We note that $\Tilde{I}_2$ is not polyconvex in $\bfF$, as previously shown by \citet{hartmann_polyconvexity_2003} and \cite{klein2022finite}. To this end, we use the modified second invariant $I_2^*$ and equivalently $K_2$, which is polyconvex in $\bfF$ (see \cite{hartmann_polyconvexity_2003} for proof).  In summary,  the inputs $(K_1,K_2,K_3)$ are polyconvex in $\bfF$. If $W_\calQ^\text{ICKAN}$ is convex and monotonically non-decreasing in $(K_1,K_2,K_3)$, then the overall strain energy density $W(\bfF)$ is polyconvex in $\bfF$ \citep{balzani2006polyconvex}. In the subsequent section, we present the \RV{monotonic} input-convex architecture for KANs to satisfy this physical constraint.
    
\end{itemize}

\subsection{\RV{Monotonic} input-convex Kolmogorov-Arnold network (ICKAN)}\label{sec:ICKAN}

\begin{figure}[t]
\centering
\includegraphics[width=1.0\textwidth]{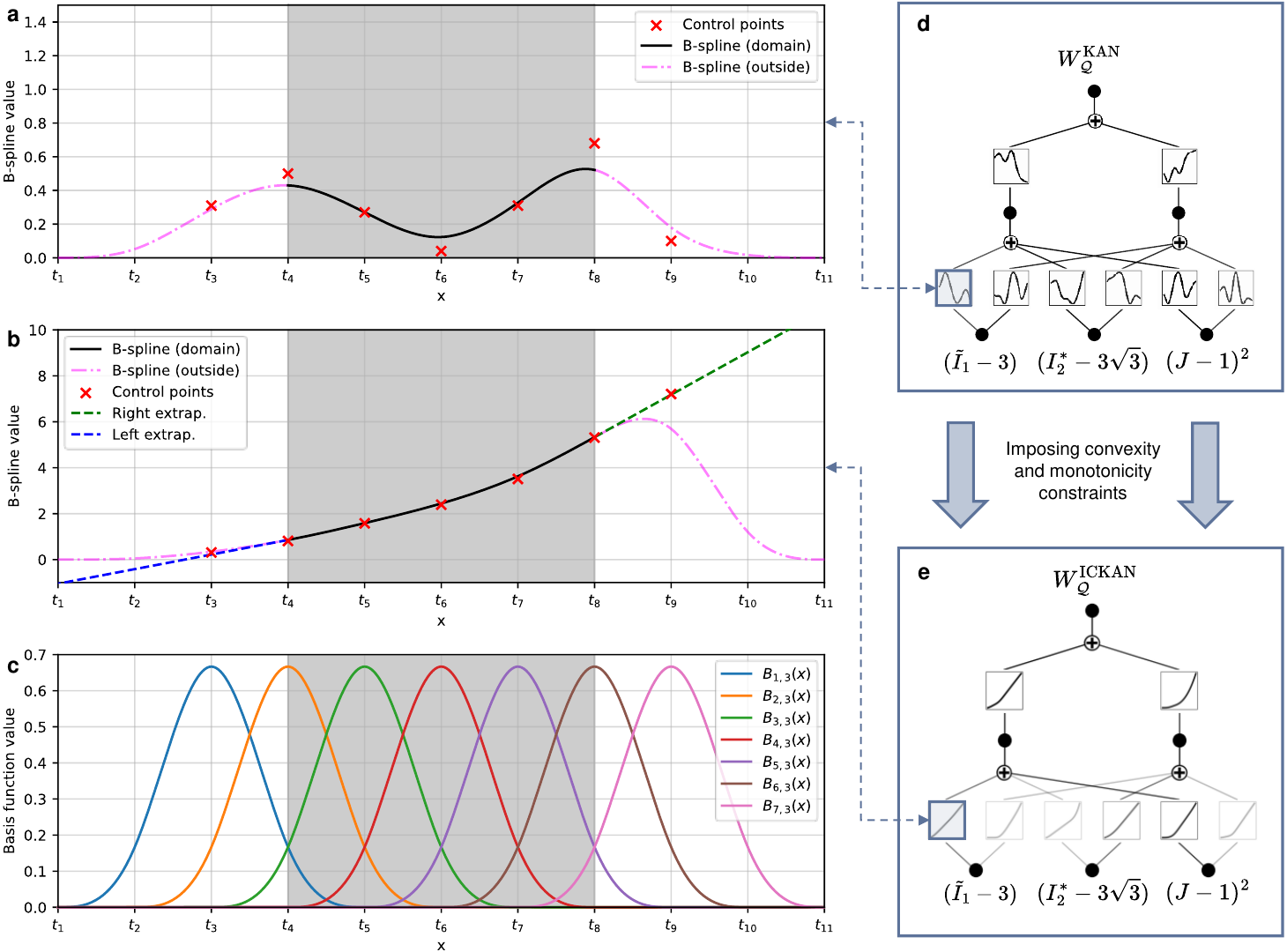}
\caption{  Imposing monotonicity and convexity constraints on the trainable B-splines of $W_{\calQ}^{\text{ICKAN}}$.  
(a) A B-spline with unconstrained control points (red), illustrating how the spline curve is constructed. The natural domain of definition for the B-spline is indicated by the shaded gray area.  
(b) The resulting B-spline curve after enforcing both monotonicity and convexity constraints, with linear extrapolation applied at both endpoints to ensure consistent behavior beyond the spline domain.  
(c) Uniform B-spline basis functions of order $k=3$.  
(d) Schematic representation of a vanilla KAN model using unconstrained trainable B-splines, where activation functions remain unrestricted.  
(e) The ICKAN model, incorporating convexity and monotonicity constraints on the trainable B-spline-based activation functions. These constraints ensure a physically consistent strain energy density. The insets at each activation node illustrate the univariate activation functions used within the respective networks.  }  

\label{fig:spline}
\end{figure}

The model architecture of $W^{\text{ICKAN}}_{\calQ}$ is based on the Kolmogorov-Arnold representation theorem, which states that any multivariate continuous function on a bounded domain can be expressed as a sum of univariate continuous functions. In its simplest form, a multivariate function $ f(x_1, \dots, x_n) $ can be written as a sum of activations of univariate functions. KANs extend this principle to deeper and wider architectures, enabling the approximation of more complex multivariate functions. \figurename~\ref{fig:spline} presents a schematic of the ICKAN architecture, which is discussed in the following sections.

We consider a multi-layer KAN with $R$ layers. For the $r^\text{th}$ layer of the KAN, let $\bfz^{(r)}\in\Rset^{n_{r}}$ be the $(n_{r})$-dimensional output. For each layer, the output is obtained as \citep{liu2024kan}
\be\label{kan-layer}
\text{KAN layer:} \qquad 
 \bfz^{(r)} = \Bigg[\sum_{j=1}^{n_{r-1}} \phi_{r-1,1,j}\Bigl(z^{(r-1)}_j\Bigr) \ , \ \dots, \ \sum_{j=1}^{n_{r-1}} \phi_{r-1,n_r,j}\Bigl(z^{(r-1)}_j\Bigr)\Bigg]^T_{n_r\times 1},
\ee
where $\bfz^{(r-1)}$ is the output of the previous layer. Here, $\{\phi_{r-1,i,j}\}_{i=1}^{n_r}$ are  $n_r$ trainable univariate functions of the $r^\text{th}$ layer transforming the $j^\text{th}$-dimension of the layer input, i.e., $z^{(r-1)}_j$. Summing all the activations $\phi_{r-1,i,j}(\cdot)$ over $j\in\{1,\dots,n_{r-1}\}$ yields the $i^\text{th}$-dimension of the layer output, i.e., $z^{(r)}_i$. The choice of the univariate activations $\phi_{r-1,i,j}(\cdot)$ can be arbitrary. However, to ensure input-convexity, special considerations apply on the choice of $\phi_{r-1,i,j}(\cdot)$. For the following discussion, we omit the subscripts on $\phi(\cdot)$ for brevity, while implicitly assuming that all $\phi$ functions are independent and have their own trainable parameters.

To ensure \RV{monotonic} input-convexity, we follow the principles that \textit{(i)} non-negative sum of convex functions are convex, \textit{(ii)}  convex non-decreasing composition of a convex function is convex \citep{boyd2004convex,amos_input_2017}, \textit{(iii)} non-negative sum of monotonically non-decreasing functions is monotonically non-decreasing, and \textit{(iv)} the composition of monotonically non-decreasing functions is monotonically non-decreasing.
\RV{A sufficient condition to ensure \RV{monotonic} input-convexity is that each KAN layer is convex and non-decreasing.} Consequently, from \eqref{kan-layer}, each univariate function $\phi$ must be convex and non-decreasing. 

We use uniform B-splines as a choice for univariate activation function, i.e.,
\begin{align}
\phi(x) &=  g(w_{s}) \psi(x)
\end{align}
for a scalar input $x$.  Here,  $w_{s}\in\Rset$ is a trainable scalar weight and $g:\Rset\rightarrow\Rset^+$ is a non-negative function. This ensures that as long as $\psi$ is convex and non-decreasing, $\phi$ remains convex and non-decreasing, while allowing for a trainable scaling parameter to alter the magnitude of the activations.  Here, we choose $g(\cdot)=\log(1+\exp(x))$, i.e., the softplus function.
The trainable univariate function $\psi(x)$ is a convex and non-decreasing uniform B-spline of order $k$ and a linear combination of $n_b$ piecewise polynomial basis functions $B_{i,k}(x)$:
\begin{align}
    \psi(x)=\sum_{i=1}^{n_b} c_i B_{i,k}(x), \qquad \text{with} \quad \sum_{i=1}^{n_b} B_{i,k}(x)=1 \quad \text{for}\quad x\in [x_\text{min},x_\text{max}].
\end{align}
Here, $c_i$ is the control point associated with the corresponding basis functions $B_{i,k}$. $[x_\text{min},x_\text{max}]$,  is natural definition domain of the B-spline curve where zeroth-order consistency is satisfied, i.e., constant functions can represented exactly. The learning process involves training the control points $\{c_i\}_{i=1}^{n_b}$ (along with the scaling weight $w_s$), for each univariate function $\phi(\cdot)$ (recall we dropped the subscripts for brevity).

To define the $k^\text{th}$-order B-spline basis functions ({illustrated in \figurename~\ref{fig:spline}c}) for each $\phi$, we consider a set of $m_b=(k+n_b+1)$ knots: $\{t_i\}_{i=1}^{m_b}$ and recursively use the algorithm introduced by \cite{de1972calculating}  as follows:

\quad Zero-order basis function ($ k=0 $):
\begin{equation}
B_{i,0}(x) =  
\begin{cases}  
1, \quad \text{if } t_i \leq x < t_{i+1}, \\  
0, \quad \text{otherwise}.  
\end{cases}  
\end{equation}  

\quad Recursive definition for higher orders ($ k > 0 $):
\begin{equation}
B_{i,k}(x) = \frac{x - t_i}{t_{i+k} - t_i} B_{i,k-1}(x) + \frac{t_{i+k+1} - x}{t_{i+k+1} - t_{i+1}} B_{i+1,k-1}(x).
\end{equation}

For the special case of a uniform B-spline, the knots are equally spaced, i.e.,
\begin{equation} 
t_{i+2} - t_{i+1} = t_{i+1} - t_{i}, \quad \forall\ i \in [1, m_b-2]. 
\end{equation}
We choose the knots by uniformly discretizing a range of pre-defined (as a hyperparameter) univariate inputs of $\phi(x)$.

For a uniform B-spline to be convex and monotonically non-decreasing, the control points must satisfy the following condition:
\begin{equation} \label{eq:convexity-condition}
c_{i+2} - c_{i+1} \geq c_{i+1} - c_{i} \geq 0, \quad \forall\  i \in [1, n_b-2]. 
\end{equation}
The proof is provided in \ref{sec:convexity proof}. We note that the convexity and non-decreasing monotonicity of a uniform B-spline curve depends entirely on the control points and is independent of the knot spacing. 
This result follows from the convex hull property of B-splines: since the B-spline is contained in the convex hull of its control points, the control points must form a convex set and be non-decreasing for the B-spline curve.
The implementation of the convexity and non-decreasing constraints are outlined in \ref{sec:constraints}.

The construction in \eqref{eq:convexity-condition} only guarantees convexity and non-decreasing monotonicity in the natural definition domain of the B-spline. The range of the knots are chosen to be sufficiently large for any reasonable data distribution relevant to hyperelasticity. Nevertheless, for extreme input data (e.g., exceptionally large strains), we extend the B-spline beyond its natural domain by maintaining a constant slope at its endpoints to preserve these properties; see \figurename~\ref{fig:spline}b for a schematic and  \ref{sec:extrapolation} for implementation details. To ensure that the spline domain  encompasses the expected input range, we set the knots' range during initialization. This update is done only once and remains fixed throughout training and inference (more details in \ref{sec:grid_init}).

In summary, the ICKAN architecture for isotropic compressible hyperelasticity can be described as follows:

\begin{subequations}
    \begin{align}
        \textbf{Input layer}: \quad & \bfF\,, \\[1mm]
        \textbf{Invariants layer}: \quad & \bfz^{(0)} = \Big[K_1=J^{-2/3}I_1-3,\ K_2=(J^{-4/3}I_2)^{3/2}-3\sqrt{3},\ K_3=(J-1)^2\Big]^T\,, \\
        \textbf{First KAN layer}: \quad & \bfz^{(1)} = \Big[\sum_{j=1}^{n_0}\phi_{0,1,j}\big(z^{(0)}_j\big) \ , \ \dots,  \ \sum_{j=1}^{n_0}\phi_{0,n_1,j}\big(z^{(0)}_j\big)\Big]^T\,, \\[1mm]
        \vdots\qquad \qquad & \notag\\
        \textbf{$r^\text{th}$ KAN layer}: \quad & \bfz^{(r)} = \Bigg[\sum_{j=1}^{n_{r-1}} \phi_{r-1,1,j}\Bigl(z^{(r-1)}_j\Bigr) \ , \ \dots, \ \sum_{j=1}^{n_{r-1}} \phi_{r-1,n_r,j}\Bigl(z^{(r-1)}_j\Bigr)\Bigg]^T\,, \\[1mm]
        \vdots\qquad \qquad & \notag\\
        \textbf{Output layer}: \quad & W^\text{ICKAN}_{\calQ} = \bfz^{(R)} = \sum_{j=1}^{n_{R-1}} \phi_{R-1,1,j}\Bigl(z^{(R-1)}_j\Bigr)\,\\
        \textbf{Trainable parameters}:\quad  & \calQ = \Big\{ \{c_k^{(r,i,j)}: k=1,\dots,n_b\},\, w_s^{(r,i,j)} : r=1,\dots,R; \; i=1,\dots,n_{r}; \; j=1,\dots,n_{r-1} \Big\}.
    \end{align}
\end{subequations}
\RV{Table~\ref{tab:parameters} provides a summary of all the architectural hyperparameters.}

\section{Unsupervised training of ICKANs}\label{sec:Problem}

\begin{figure}[t]
\centering
\includegraphics[width=\textwidth]{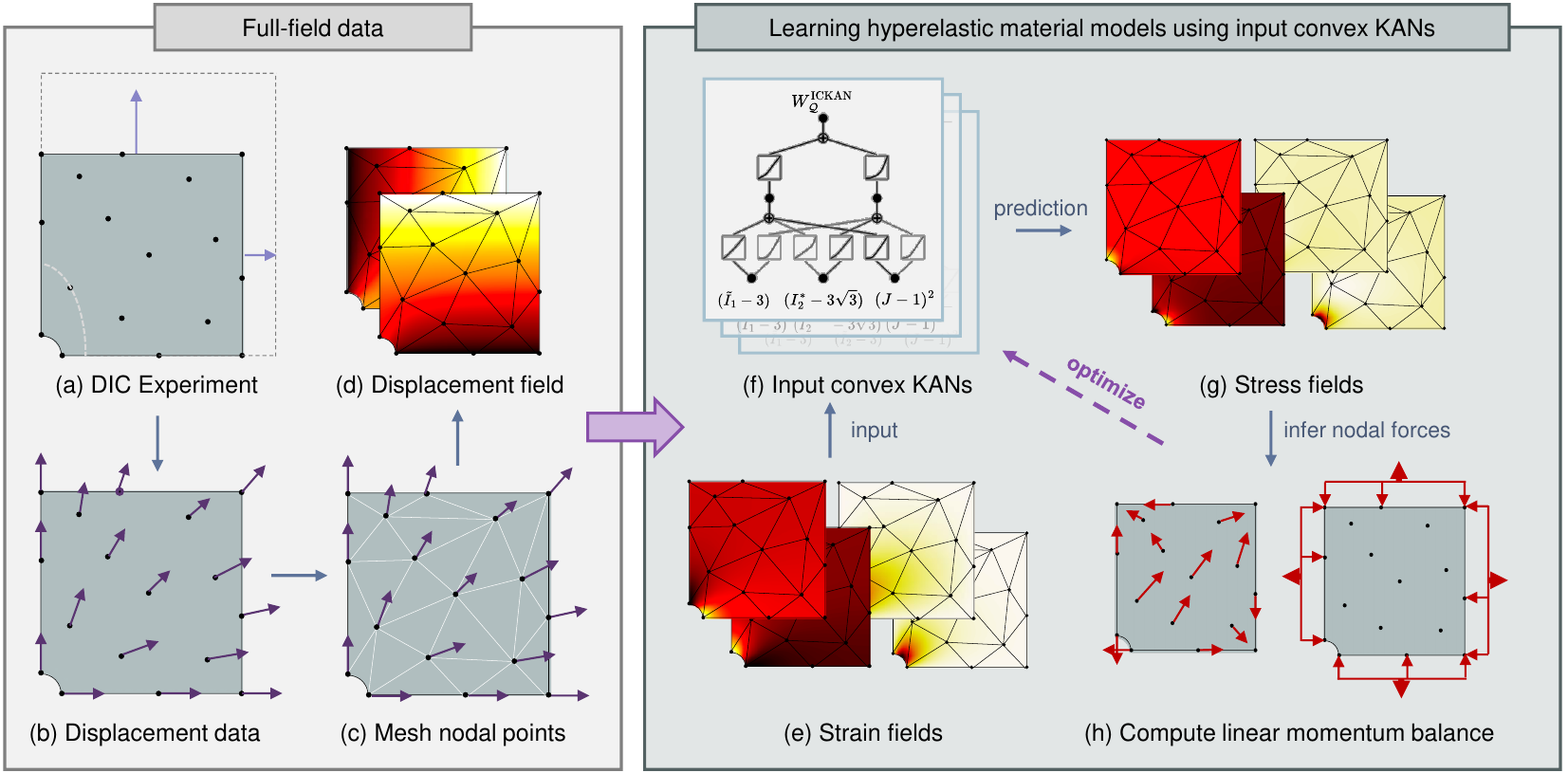}
\caption{Schematic overview of the unsupervised deep learning approach for hyperelastic constitutive modeling using ICKANs. In (a) and (b), pointwise measurements of displacements and reaction forces are acquired from a hyperelastic specimen under quasi-static deformation using a DIC setup. These data, together with a finite element mesh of the domain (c), are used to reconstruct continuous displacement and strain fields (d,e). The physics-consistent ICKAN-based constitutive model (f) then predicts the corresponding stress fields at the quadrature points of each element (g) given the strains. The computed stress fields serve to calculate both the internal and external nodal forces (h). By employing the weak form of the conservation of linear momentum, the residual forces are minimized—applied pointwise for free degrees of freedom and in an aggregated manner for fixed degrees of freedom corresponding to each set of Dirichlet constraints with measured reaction forces. This iterative optimization process refines the parameters of the ICKAN-based constitutive models.}
\label{fig:overview}
\end{figure}

Instead of training the ICKAN model on labeled stress-strain pairs, we train the model on full-field displacement/strain-fields and global force data which are realistically available through mechanical testing and DIC techniques. To achieve this, we extend our previously introduced NN-based constitutive modeling framework, NN-EUCLID \citep{thakolkaran2022nn}, to ICKANs. Here, we provide a brief review of NN-EUCLID in the context of ICKANs. \figurename~\ref{fig:overview} illustrates a schematic overview of the ICKAN and its unsupervised training framework.

Consider a hyperelastic specimen undergoing quasi-static deformation in a two-dimensional reference domain $\Omega \subset \mathbb{R}^2$. The specimen is designed with a complex geometry (for example, a plate with a hole) to induce diverse and heterogeneous strain states, while the material itself is assumed to be homogeneous and isotropic. Without loss of generality, we assume plane strain conditions.

Boundary conditions are applied such that Dirichlet conditions are enforced on $\partial \Omega_u \subset \partial \Omega$ and Neumann conditions on the remainder, $\partial \Omega_t = \partial \Omega \setminus \partial \Omega_u$. For simplicity, our analysis focuses on displacement-controlled loading (i.e., Dirichlet boundary conditions), while noting that applied forces in load-controlled scenarios are equivalent to reaction forces under displacement control. The dataset comprises of $n_t$ snapshots of displacement measurements,
\be
\mathcal{U} = \left\{ \bfu^{a,t} \in \mathbb{R}^2 : a=1,\dots,n_n;\ t=1,\dots,n_t \right\},
\ee
recorded at $n_n$ reference points
\be
\mathcal{X} = \left\{ \boldsymbol{X}^a \in \Omega : a=1,\dots,n_n \right\}.
\ee
Additionally, for each snapshot, $n_\beta$ reaction forces 
\be
\left\{ R^{\beta,t} : \beta=1,\dots,n_\beta;\ t=1,\dots,n_t \right\}
\ee
are measured at selected Dirichlet boundaries (e.g., using load cells). For brevity, the superscript $(\cdot)^t$ is omitted in the subsequent discussion, although the numerical procedure is applied independently to every snapshot.

Given these limited measurements, the primary goal is to infer the constitutive model $W(\bfF)$ governing the stress–strain response.

To reconstruct the displacement field, the reference domain $\mathcal{X}$ is discretized using linear triangular elements (each with a single quadrature point at its barycenter), yielding the approximation
\be\label{disp-field-approx}
\boldsymbol{u}(\boldsymbol{X}) = \sum_{a=1}^{n_n}N^a(\boldsymbol{X})\ \boldsymbol{u}^{a}.
\ee
Here, $N^a : \Omega \to \mathbb{R}$ represents the shape function associated with node $\boldsymbol{X}^a$. The corresponding deformation gradient field is approximated as
\begin{align}
\boldsymbol{F}(\boldsymbol{X}) = \boldsymbol{I} + \sum_{a=1}^{n_n}\boldsymbol{u}^{a}\otimes \nabla N^a(\boldsymbol{X}),
\end{align}
where $\boldsymbol{I}$ is the identity matrix, and $\nabla$ denotes the gradient operator with respect to the reference coordinates.

As outlined in  \cite{flaschel_unsupervised_2021} and \cite{thakolkaran2022nn}, we leverage the conservation of linear momentum to guide the learning of the constitutive model, which eliminates the need of stress labels. Assuming quasi-static loading and negligible body forces, the weak form of the linear momentum balance in the reference domain $\Omega$ is given by
\begin{equation}\label{weak-form}
    \int_{\Omega} \boldsymbol{P} : \nabla\boldsymbol{v}\, \mathrm{d}V - \int_{\partial\Omega_t} \hat{\boldsymbol{t}} \cdot \boldsymbol{v}\, \mathrm{d}S = 0 \quad \forall \ \text{admissible } \boldsymbol{v}\,,
\end{equation}
where $\hat{\boldsymbol{t}}$ denotes the prescribed tractions and $\boldsymbol{v}$ is a test function that vanishes on the Dirichlet boundary $\partial\Omega_u$. The weak formulation is preferred over the strong form, as it avoids the need for second derivatives, which are sensitive to noise.

Let $\mathcal{D} = \{(a,i): a=1,\dots,n_n;\ i=1,2\}$ denote the set of displacement degrees of freedom, which we partition into:
\begin{itemize}
    \item $\mathcal{D}^\text{free}$: unconstrained degrees of freedom,
    \item $\mathcal{D}^\text{fix}_\beta$ (with $\beta=1,\dots,n_\beta$): degrees of freedom under Dirichlet constraints that contribute to the observed reaction force $R^\beta$.
\end{itemize}

Approximating the test function by
\be
\boldsymbol{v}(\boldsymbol{X}) = \sum_{a=1}^{n_n} N^a(\boldsymbol{X})\,\boldsymbol{v}^a,\qquad \text{with} \quad  v_i^a = 0 \quad \forall\ (a,i)\in\bigcup_{\beta=1}^{n_\beta} \mathcal{D}^\text{fix}_\beta\,,
\ee
the weak form \eqref{weak-form} reduces to
\begin{equation}\label{reducedWeakForm}
    \sum_{a=1}^{n_n} v_i^a f_i^a = 0\,, \quad \text{with} \quad f_i^a = \underbrace{\int_{\Omega} P_{ij}\,\nabla_j N^a\,\mathrm{d}V}_{\text{internal force}} - \underbrace{\int_{\partial\Omega_t} \hat{t}_i N^a\,\mathrm{d}S}_{\text{external force}}\,.
\end{equation}
Here, the integrals are computed using numerical quadrature over the same discretization and mesh as \eqref{disp-field-approx}. 
Since the test functions are arbitrary, the force residual must vanish at every free degree of freedom,
\begin{equation}\label{free-constraints}
    f_i^a = 0 \quad \forall\ (a,i) \in \mathcal{D}^\text{free}\,.
\end{equation}

{At the fixed degrees of freedom, the internal and external forces are counteracted by the reaction force imposed by the Dirichlet constraints. Since point-wise reaction forces cannot be measured experimentally, they are assumed to be unavailable. Instead, only the total reaction forces integrated over the boundary segments are known. As a result, the global force balance for each measured reaction force is expressed as  
\be\label{fix-constraints}
\sum_{(a,i)\in\calD^\text{fix}_\beta} f^a_i = R^\beta \qquad \forall \qquad \beta=1,\dots,n_\beta,
\ee  
where the summation is carried out over all point-wise forces associated with the degrees of freedom in the $\beta^\text{th}$ Dirichlet constraint, denoted as $\calD^\text{fix}_\beta$. As noted earlier, the superscript $(\cdot)^t$ has been omitted for brevity, but the above force balance conditions apply to all data snapshots at $t=1,\dots,n_t$.  }

Our goal is {then} to learn the constitutive model $W(\bfF)$ in \eqref{eq:ansatz} (with $\bfP(\bfF)$ given by \eqref{eq:PK_ansatz}), now parameterized by the \RV{monotonic} input-convex Kolmogorov–Arnold network $W^\text{ICKAN}_{\calQ}\,,$
such that the displacement and reaction force data satisfy the physics-based constraints \eqref{free-constraints} and \eqref{fix-constraints}. We formulate the inverse problem as the minimization of the force balance residuals \citep{thakolkaran2022nn} with respect to the trainable parameters of the ICKAN:
\begin{equation}\label{loss}
    \calQ \leftarrow \arg\min_{\calQ} \sum_{t=1}^{n_t} \Bigg[
    \underbrace{\sum_{(a,i)\ \in \ \mathcal{D}^\text{free}} \left(f_i^{a,t}\right)^2}_{\text{free degrees of freedom}}
    + \underbrace{\sum_{\beta=1}^{n_\beta} \Bigl(R^{\beta,t} - \sum_{(a,i)\ \in\ \mathcal{D}^\text{fix}_\beta} f_i^{a,t}\Bigr)^2}_{\text{fixed degrees of freedom}}
    \Bigg]\,.
\end{equation}
This optimization is performed via gradient-based minimization, enabled by automatic differentiation \RV{(see Table \ref{tab:parameters} for details on the optimizer and related hyperparameters)}.

\section{Results}\label{sec:Results}
\subsection{Numerical benchmarks}
\begin{figure}[t]
\centering
\includegraphics[width=0.7\textwidth]{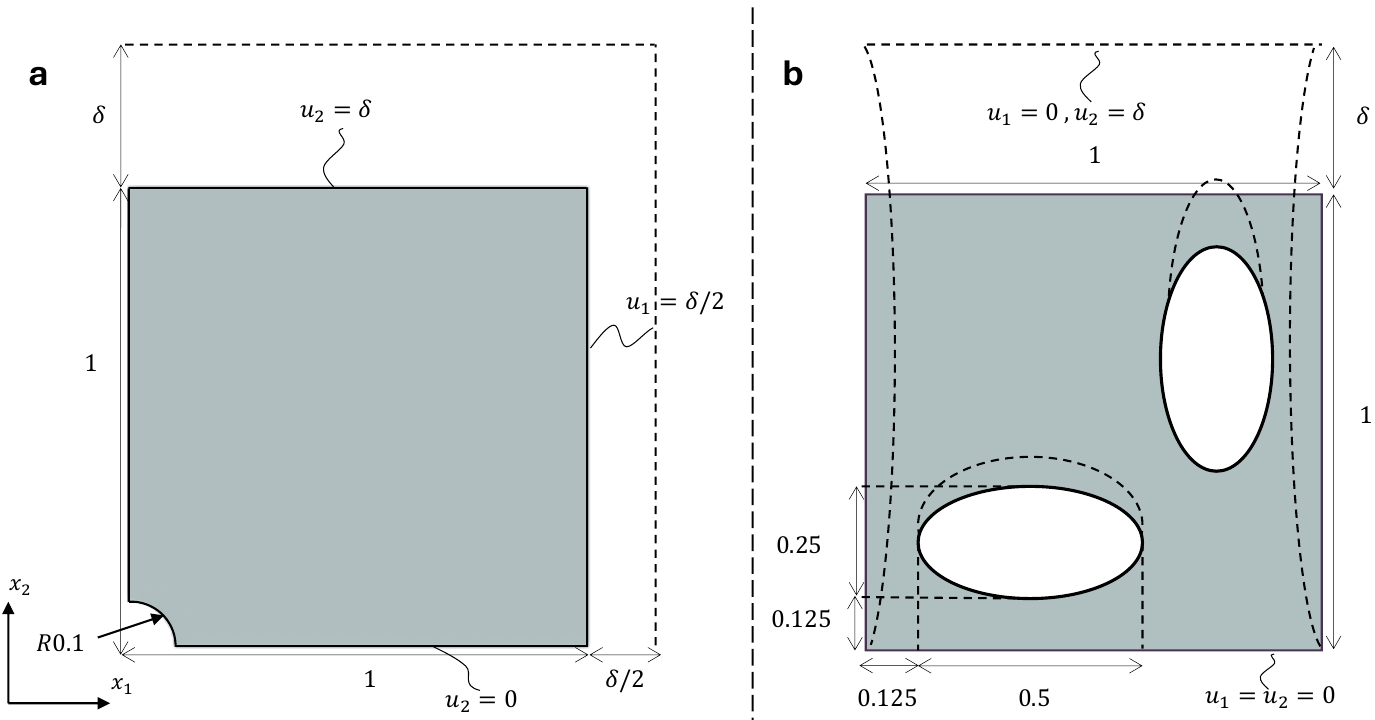}
\caption{Figures adapted from \citep{thakolkaran2022nn}. 
(a) \textit{Training specimen}: A square plate with a hole in the bottom-left corner is subjected to displacement-controlled asymmetric biaxial tension. The resulting dataset (noisy full-field displacements and reaction forces) is used to train the ICKAN-based constitutive models. 
(b) \textit{Validation specimen}: A square plate containing two asymmetric elliptical holes is subjected to displacement-controlled uniaxial tension. This geometry is employed solely for validation, and no data from these experiments enter the training process. All lengths and displacements are normalized by the side length of the undeformed specimen. \RV{The parameters used to generate the data can be found in \tablename~\ref{tab:parameters}}.}
\label{fig:geometries}
\end{figure}

To evaluate the ICKAN-based constitutive models, we use the dataset provided by \citep{thakolkaran2022nn}, which emulates a digital image correlation (DIC) experiment by simulating a hyperelastic square plate with a hole under plane-strain conditions using the finite element method (FEM); see \figurename~\ref{fig:geometries}a for schematic. The specimen is subjected to displacement-controlled asymmetric biaxial tension, with symmetry enforced along the bottom and left boundaries and loading parameter $\delta$.  This configuration yields diverse and heterogeneous strain states, making the data ideal for training a generalizable constitutive model from a single experiment; hence, the specimen is termed the \textit{training specimen}.

The dataset includes synthetic data corresponding to several well-established material models:

\begin{enumerate} 
\item Neo-Hookean (NH) model: 
\begin{equation}\label{eq:NH2}
W(\bfF) = 0.5 (\Tilde{I}_1-3) + 1.5 (J-1)^2. 
\end{equation} 

\item Isihara (IH) model  \citep{isihara_statistical_1951}: 
\begin{equation}\label{eq:IH}
W(\bfF) = 0.5 (\Tilde{I}_1-3) + (\Tilde{I}_2-3) + (\Tilde{I}_1-3)^2 + 1.5 (J-1)^2. 
\end{equation} 

\item Haines-Wilson (HW) model  \citep{haines_strain-energy_1979}: \begin{equation}\label{eq:HW}
W(\bfF) = 0.5 (\Tilde{I}_1-3) + (\Tilde{I}_2-3) + 0.7 (\Tilde{I}_1-3)(\Tilde{I}_2-3) + 0.2 (\Tilde{I}_1-3)^3 + 1.5 (J-1)^2. 
\end{equation} 

\item Gent-Thomas (GT) model  \citep{gent_forms_1958}: \begin{equation}\label{eq:GT}
W(\bfF) = 0.5 (\Tilde{I}_1-3) + \log(\Tilde{I}_2/3) + 1.5 (J-1)^2. 
\end{equation} 

\item Arruda-Boyce (AB) model  \citep{arruda_three-dimensional_1993}: \begin{equation}\label{eq:AB} 
W(\bfF) = 2.5 \sqrt{N_c}\left[ \beta_c \lambda_c - \sqrt{N_c}\log \left(\frac{\sinh \beta_c}{\beta_c}\right)\right] - c_{\text{AB}} + 1.5 (J-1)^2, \end{equation} 
where $\lambda_c=\sqrt{{\Tilde{I}_1}/{3}}$, $\beta_c=\mathcal{L}^{-1}\left({\lambda_c}/{\sqrt{N_c}}\right)$, and $\mathcal{L}^{-1}$ denotes the inverse Langevin function. Here, $N_c=28$ represents the number of polymeric chain segments, and $c_{\text{AB}}\approx 3.7910$ offsets the energy density to zero at $\bfF=\bfI$, since the Arruda-Boyce formulation does not inherently satisfy this condition. 

\item Ogden (OG) model  \citep{ogden_large_1972}: \begin{equation}\label{eq:feature_OG}
\RV{W(\bfF)} = \frac{\mu}{\eta}\left( \Tilde{\lambda}_1^{\eta}+\Tilde{\lambda}_2^{\eta}+\Tilde{\lambda}_3^{\eta}-3\right) + 1.5 (J-1)^2, \quad \text{with} \quad \Tilde{\lambda}_k = J^{-1/3}\lambda_k,\ k=1,2,3,
\end{equation} 
where $\lambda_1,\lambda_2,\lambda_3$ are the principal stretches and $\mu=\eta=1.3$. 
\end{enumerate}

\RV{Note that the models given in equations~\eqref{eq:IH}--\eqref{eq:GT} are not polyconvex, as they involve the modified invariant $\Tilde{I}_2$, which does not fulfill ellipticity. For discussions on the polyconvexity of the Ogden model, we refer the reader to the work of \citet{hartmann_polyconvexity_2003}.
These models are included as they are commonly used in literature and provide a representative benchmark for evaluating constitutive modeling approaches. Additionally, without loss of generality, all physical quantities in the material models are considered as non-dimensionalized.}

For benchmarking purposes, the ICKAN-based model is trained to approximate these material laws using the NN-EUCLID loss described in \eqref{loss}. Additional details regarding hyperparameters and the training algorithm are provided in~\ref{sec:training-details}.

Considering that real-world DIC measurements contain noise, artificial noise has been added to the displacement data. The noise level, determined by the pixel resolution of the imaging system, remains constant for each degree of freedom across all load steps. Specifically, the displacement data is modeled as \begin{align} u_i^{a,t} = u_i^{\text{fem},a,t} + \eps_i^{a,t}, \quad \text{with} \quad \eps_i^{a,t} \sim \mathcal{N}(0,\sigma_u^2) \quad \forall \ (a,i) \in \mathcal{D},\ t \in {1,\dots,n_t}. \end{align} Here, $u_i^{\text{fem},a,t}$ represents the FEM-computed displacement, while $\eps_i^{a,t}$ is sampled from a normal distribution with zero mean and standard deviation $\sigma_u$. Following  \cite{flaschel_unsupervised_2021, thakolkaran2022nn, joshi_Bayesian}, two noise levels are considered (normalized by specimen length): $\sigma_u=10^{-4}$ (low noise) and $\sigma_u=10^{-3}$ (high noise), which emulate the noise levels representative of modern DIC setups.

\RV{We use pre-processed synthetic datasets from \citet{thakolkaran2022nn}, where the displacement fields were already denoised following standard DIC post-processing routines. As described in that work, kernel ridge regression (KRR) was used for spatial denoising. To enhance data efficiency, the denoised displacements on the high-resolution mesh with $63{,}601$ nodes are projected onto a coarser mesh with $n_n = 1441$ nodes, which serve as the training dataset for the ICKAN models.}

\subsection{Accuracy and generalizability beyond the training data}

We train $n_e$ ICKAN models with identical architectures, each trained independently with different random initializations to mitigate bad local minima in the optimized model parameters. We choose the model whose final loss (as defined in \eqref{loss}) is the lowest loss in the ensemble; the others are discarded.

\begin{figure}[t]
		\centering
		\includegraphics[width=1.0\textwidth]{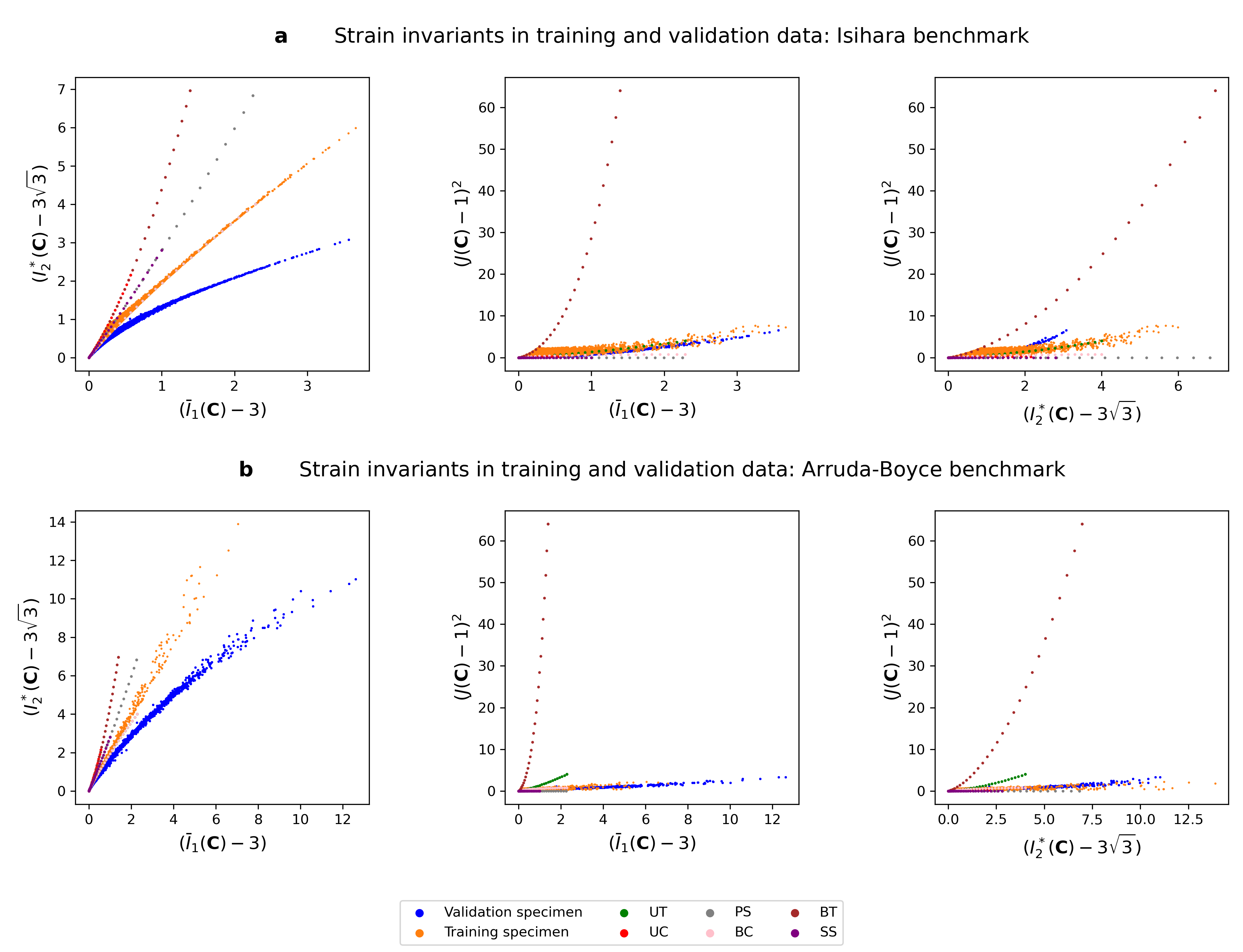}
		\caption{Two-dimensional projections of the strain invariants $(\Tilde{I}_1-3)$, $(I_2^*-3\sqrt{3})$, and $(J-1)^2$ for all elements in the training specimen (\figurename~\ref{fig:geometries}a) across all load steps, shown for (a) Arruda-Boyce \eqref{eq:AB} and (b) Isihara \eqref{eq:IH} benchmarks. The figure also includes the strain invariants of the validation specimen (\figurename~\ref{fig:geometries}b) across all elements and load steps, along with those corresponding to the six evaluation deformation paths \eqref{eq:strain_paths}. All data presented are derived from the ground-truth model without noise, including the strain invariants of the validation specimen. }\label{fig:invariants-space_AB_IH}
\end{figure}

For each benchmark problem (Equations \eqref{eq:NH2}--\eqref{eq:feature_OG}), the performance of the ICKAN-based constitutive models is evaluated against the corresponding ground truth along six deformation paths:
\be
\begin{aligned}
\boldsymbol{F}^{\text{UT}}(\gamma) = \begin{bmatrix}
1 + \gamma & 0 & 0\\
0 & 1 & 0 \\
0 & 0 & 1
\end{bmatrix}, \ \ \
\boldsymbol{F}^{\text{UC}}(\gamma) = \begin{bmatrix}
\frac{1}{1 + \gamma} & 0 & 0\\
0 & 1& 0 \\
0 & 0 & 1
\end{bmatrix}, \ \ \ 
\boldsymbol{F}^{\text{BT}}(\gamma) = \begin{bmatrix}
1 + \gamma & 0& 0\\
0 & 1+ \gamma& 0 \\
0 & 0 & 1
\end{bmatrix}, \\
\boldsymbol{F}^{\text{BC}}(\gamma) = \begin{bmatrix}
\frac{1}{1 + \gamma} & 0& 0 \\
0 & \frac{1}{1 + \gamma}& 0 \\
0 & 0 & 1
\end{bmatrix}, \ \ \
\boldsymbol{F}^{\text{SS}}(\gamma) = \begin{bmatrix}
1 & \gamma& 0\\
0 & 1& 0 \\
0 & 0 & 1
\end{bmatrix}, \ \ \
\boldsymbol{F}^{\text{PS}}(\gamma) = \begin{bmatrix}
1 + \gamma & 0& 0\\
0 & \frac{1}{1 + \gamma}& 0 \\
0 & 0 & 1
\end{bmatrix},
\end{aligned}
\label{eq:strain_paths}
\ee
where \RV{$ \gamma \in [0,2] $ denotes the loading parameter for UT, BT, and BC, and $ \gamma \in [0,1] $ for UC, SS, and PS (as these three cases already involve significant extrapolation; as shown in \figurename~\ref{fig:invariants-space_AB_IH})}. These deformation paths are \textit{used solely for evaluation} and not for training. The abbreviations used are as follows --- UT: uniaxial tension, UC: uniaxial compression, BT: biaxial tension, BC: biaxial compression, SS: simple shear, PS: pure shear. \RV{Note that, while the uniaxial tension load case represents a uniaxial strain state, it induces a multiaxial stress state.}

\figurename s~\ref{fig:NH2_IH_HW_noise=low_W}-\ref{fig:GT_AB_OG_noise=low_W} and \ref{fig:NH2_IH_HW_noise=low_Pij}-\ref{fig:GT_AB_OG_noise=low_Pij} (low noise) and \figurename s~\ref{fig:NH2_IH_HW_noise=high_W}-\ref{fig:GT_AB_OG_noise=high_W} and \ref{fig:NH2_IH_HW_noise=high_Pij}-\ref{fig:GT_AB_OG_noise=high_Pij} (high noise) confirm that both the ICKAN-based strain energy density as well as corresponding stress predictions closely match the ground truth across all benchmarks, deformation paths, and noise levels. This verifies the generalizability of the trained constitutive models.  \RV{We observe that in some cases the predictions start to deteriorate at larger strains. However, the issue can be addressed by conducting additional experiments (and include them as training data) or using specimens with a larger diversity in strain states to improve generalizability.}

We highlight that the \RV{monotonic} input-convex architecture of KANs \RV{guarantees} learning physically admissible models. In \ref{sec:nonconvex}, we demonstrate the performance of vanilla KANs without \RV{monotonicity and} input-convexity. We observe poor prediction accuracies for both strain energy densities and stresses relative to their \RV{monotonic} input-convex variants. In addition, we observe non-physical behaviors such as negative stiffness and fictitious material instabilities without \RV{monotonicity and} input-convexity in the KAN architecture.  

\begin{figure}[t]

\centering
\includegraphics[width=1.0\textwidth]{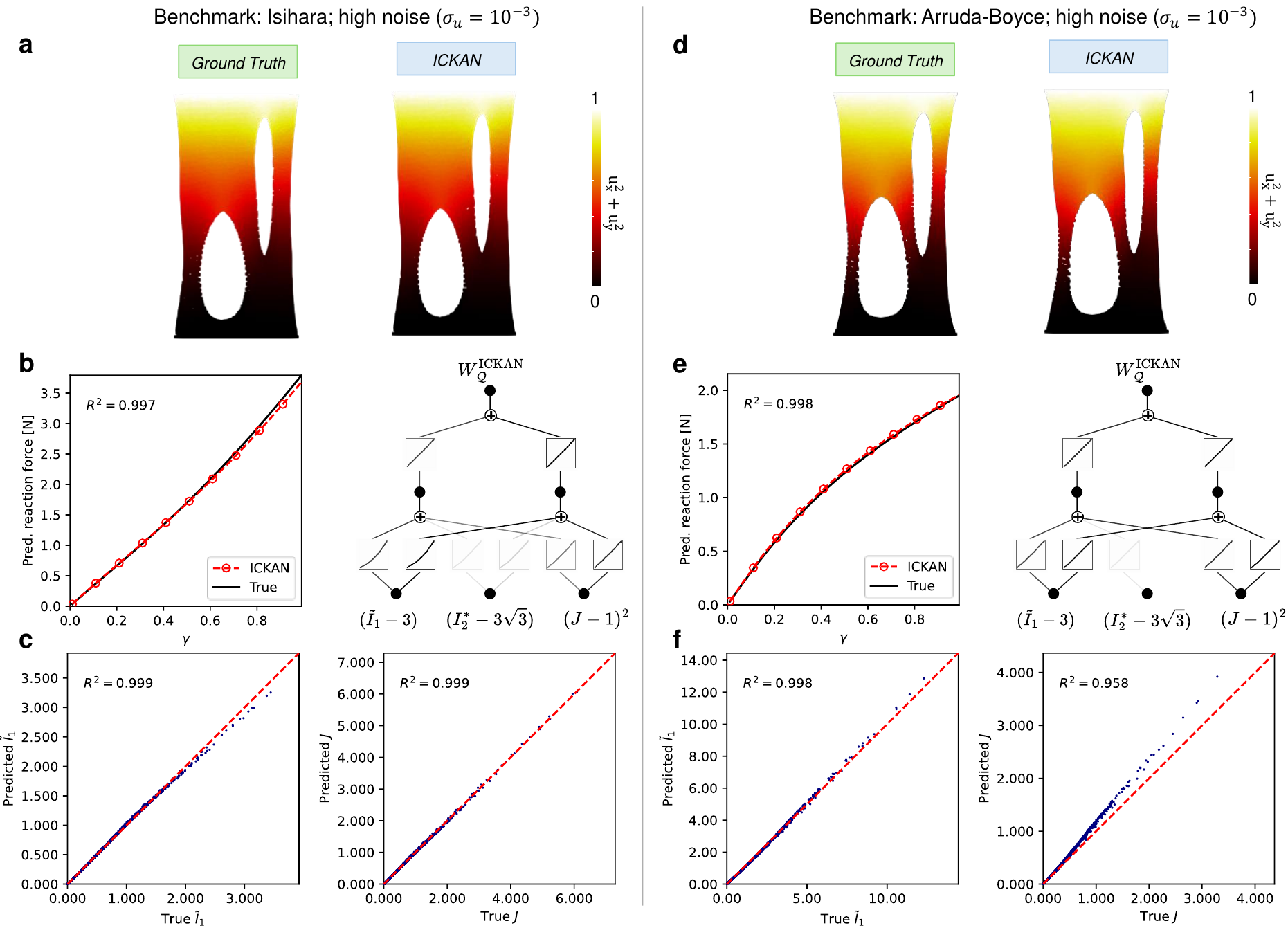}
\caption{
Comparison of FEM for the validation specimen (see Fig. \ref{fig:geometries}b) between the ground-truth and ICKAN-based constitutive model for the Isihara benchmark \eqref{eq:IH} (a-c) and the Arruda-Boyce benchmark \eqref{eq:AB} (d-f) (trained on high noise data). (a \& d) Deformed shape of the validation specimen, with colors indicating displacement magnitude, as obtained from both the ground-truth (left) and ICKAN-based (right) simulations. (b \& e) Comparison of the reaction force on the top surface plotted against the loading parameter $\gamma$ (left) and the trained ICKAN-model illustrated with the in-built plotting functionality (right) (c \& f) Parity plot of the first strain invariant $\tilde{I}_1$ (left) and the determinant $J$ (right) predicted by the ICKAN model versus the ground-truth values, with the gray dashed line representing perfect agreement (zero intercept, unit slope) and the $R^2$ value indicating accuracy.
}
\label{fig:kan-validation-AB-IH}
\end{figure}

We further validate the ICKAN-based constitutive models by deploying them within a finite element simulation framework. Linear triangular elements and a Newton-Raphson nonlinear solver are employed, with element-level stress and tangent modulus computed via automatic differentiation from Equations \eqref{eq:PK_ansatz} and \eqref{eq:tangent}.  \RV{Since tangent stiffness calculations require the energy density to be at least twice differentiable, the splines must have a minimum order of $k=3$ to ensure adequate smoothness. In this work, we use splines of order $k=5$.} \RV{These simulations were carried out using a custom in-house finite element solver implemented in Python.} To assess generalization beyond the training domain, a validation specimen with a more complex geometry—featuring two asymmetric elliptical holes and subjected to quasi-static uniaxial loading—is considered (see \figurename~\ref{fig:geometries}b). The lowest-loss ICKAN models (for each material model, respectively), selected from the ensemble as described above and trained on high-noise data, are used in these simulations. As shown in \figurename~\ref{fig:kan-validation-AB-IH}, the finite element solutions for two representative ground-truth models (Isihara \eqref{eq:IH} and Arruda-Boyce \eqref{eq:AB}, respectively) exhibit excellent agreement with those obtained using the ICKAN-based models. \RV{This is a strong indication that the ICKAN approach is not overfitting to the training data but instead captures the underlying constitutive behavior in a generalizable form.} Quantitatively, this is supported by high goodness-of-fit $R^2$ scores: greater than 0.99 for Isihara and greater than 0.95 for Arruda-Boyce in the high-noise case, for element-wise strain invariants and close matching of reaction forces. Remarkably, the strain states of this validation specimen are completely different from that of the training specimen -- as shown in \figurename~\ref{fig:invariants-space_AB_IH}; yet the ICKANs show excellent generalization capability to these unseen strain states. \RV{In~\ref{sec:euclid}, we also provide a comparative analysis of the ICKAN framework with the EUCLID approach of \cite{flaschel_unsupervised_2021}.}

\subsection{Interpretability}

One advantage of ICKANs over ICNNs and other black-box neural networks is the enhanced interpretability they offer in constitutive modeling. This interpretability manifests in two key aspects.

\textit{(i)} Due to the highly nonlinear and expressive nature of the activations---enabled by trainable splines---only a few layers with a small number of dimensions are required in KANs. Consequently, each activation function in ICKANs can be plotted and interpreted individually; see the insets of \figurename~\ref{fig:spline}d and e. In contrast, ICNNs and similar classical models rely on less expressive nonlinear activations (e.g., rectified linear unit or softplus), which necessitate many more linear layers with higher dimensionality to achieve comparable performance. This results in a significantly larger number of parameters and reduced interpretability.

\textit{(ii) } The Kolmogorov-Arnold representation allows the network architecture to be decomposed into univariate activation functions. This decomposition enables the extraction of analytical expressions at each activation node using symbolic regression, which can then be sequentially assembled across the entire network to form an analytical expression for the entire constitutive model. Such a univariate decomposition is not possible with ICNNs and similar models, thereby limiting their capacity for symbolic interpretation.

To quantitatively demonstrate the latter aspect, we extract the symbolic expression of the strain energy density from the trained ICKAN model via symbolic regression at each activation node of the network independently. These expressions are derived using a small library of univariate symbolic functions: \RV{$\{x, \exp(x), \log (1+\exp(x)), \log (1+\exp(x))^2, \log (1+\exp(x))^3, \log (1+\exp(x))^4\}$}. Each candidate function in the library is purposefully chosen to be convex and non-decreasing. In \ref{sec:symbolic_regression_details}, we introduce a \RV{monotonic} input-convex symbolic regression strategy that ensures that the symbolic approximation of the general ICKAN-based hyperelastic model remains polyconvex.

\begin{figure}
\centering
\text{Benchmark: Strain energy density predictions, low noise ($\sigma_u=10^{-4}$)}
\includegraphics[width=0.9\textwidth]{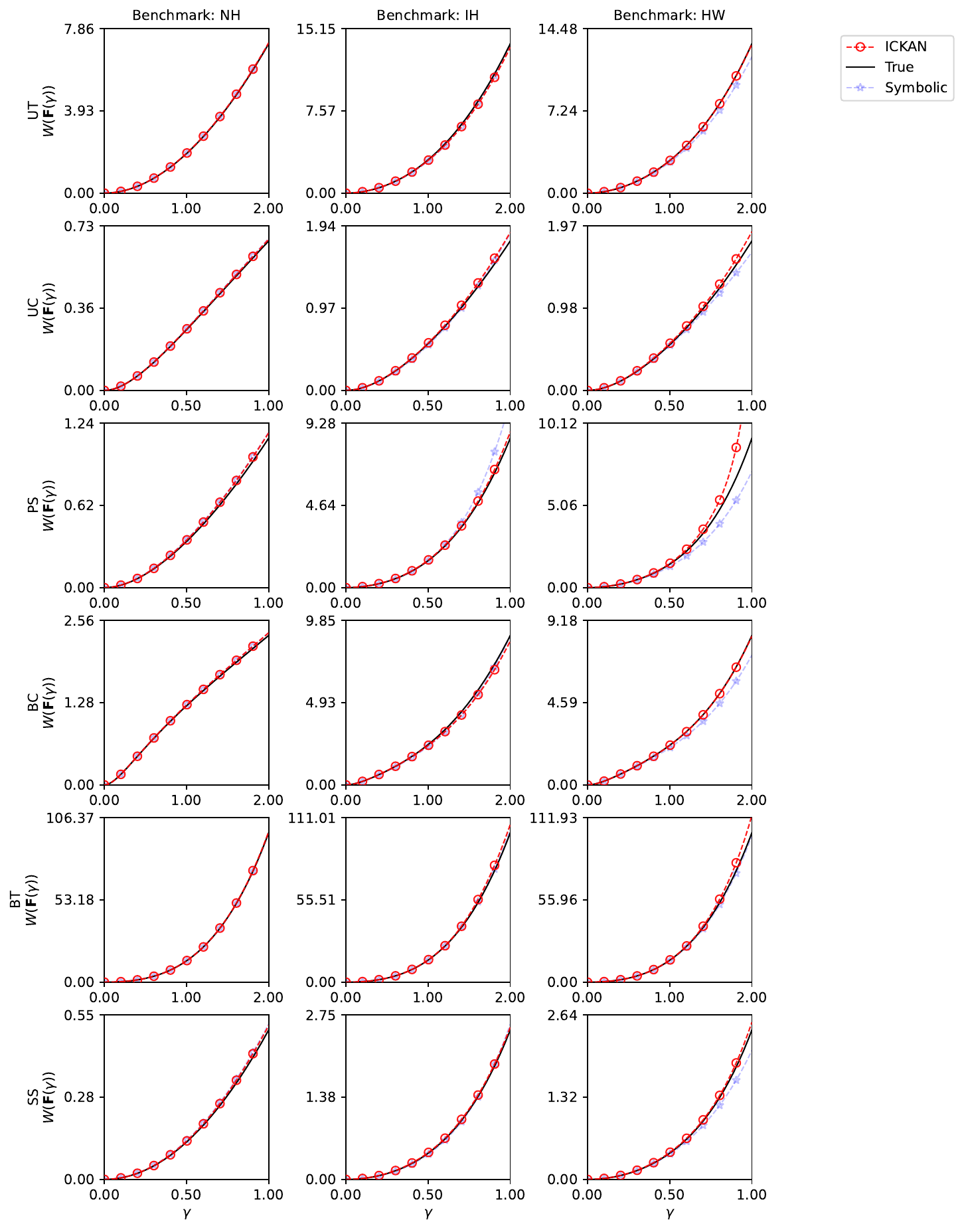}
\caption{Strain energy density $W(\bfF(\gamma))$ predictions for the deformation paths in \eqref{eq:strain_paths}, evaluated under low noise $(\sigma_u = 10^{-4})$. Results are shown for the best ICKAN-based constitutive model in the ensemble, the model obtained through symbolic regression of the ICKAN, as well as for the hidden true model, for the NH \eqref{eq:NH2}, IH \eqref{eq:IH}, and HW \eqref{eq:HW} benchmarks.}
\label{fig:NH2_IH_HW_noise=low_W}
\end{figure}

\begin{figure}
\centering
\text{Benchmark: Strain energy density predictions, low noise ($\sigma_u=10^{-4}$)}
\includegraphics[width=0.9\textwidth]{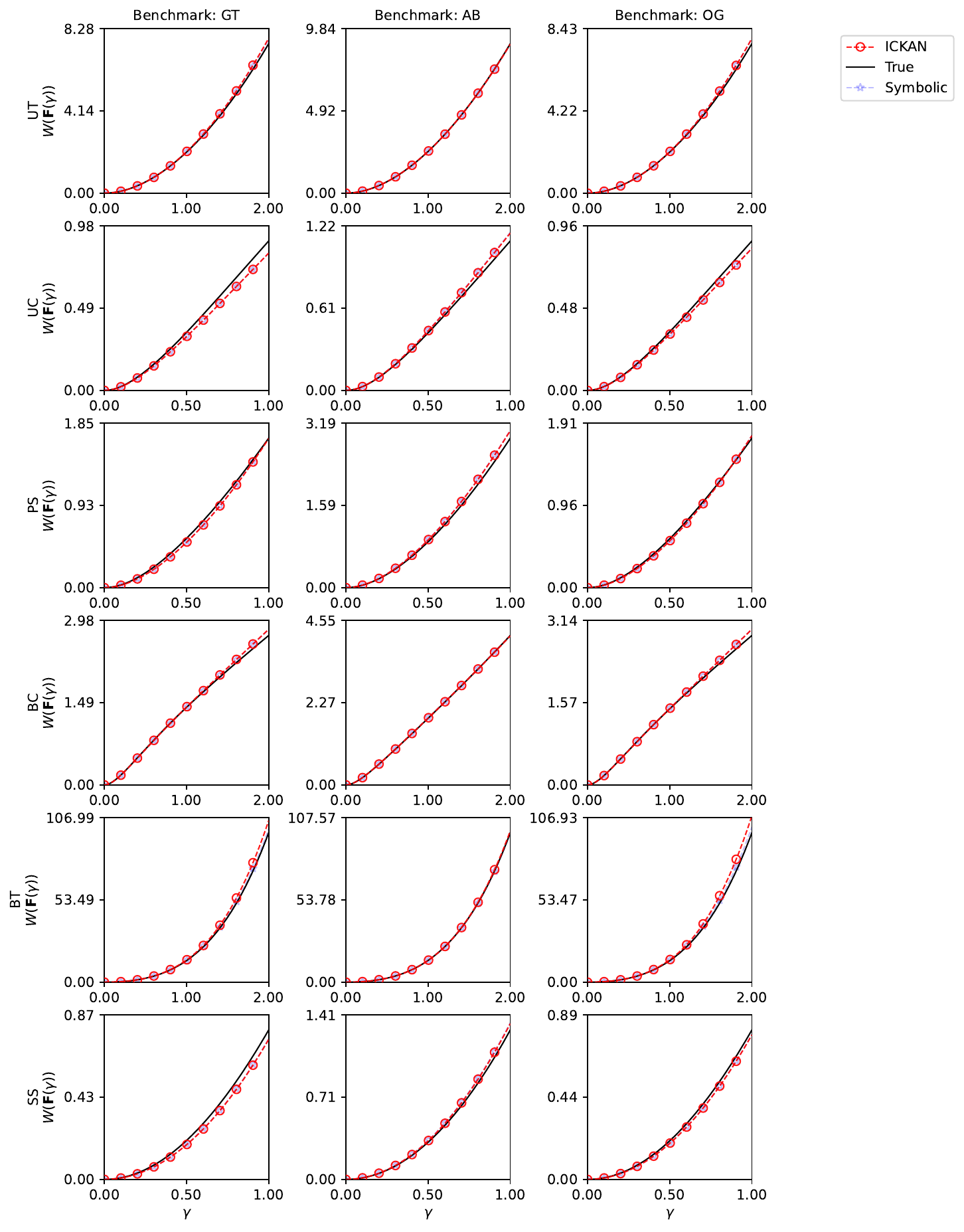}
\caption{Strain energy density $W(\bfF(\gamma))$ predictions for the deformation paths in \eqref{eq:strain_paths}, evaluated under low noise $(\sigma_u = 10^{-4})$. Results are shown for the best ICKAN-based constitutive model in the ensemble, the model obtained through symbolic regression of the ICKAN, as well as for the hidden true model, for the GT \eqref{eq:GT}, AB \eqref{eq:AB}, and OG \eqref{eq:feature_OG} benchmarks.}
\label{fig:GT_AB_OG_noise=low_W}
\end{figure}

\begin{figure}
\centering
\text{Benchmark: Strain energy density predictions, high noise ($\sigma_u=10^{-3}$)}
\includegraphics[width=0.9\textwidth]{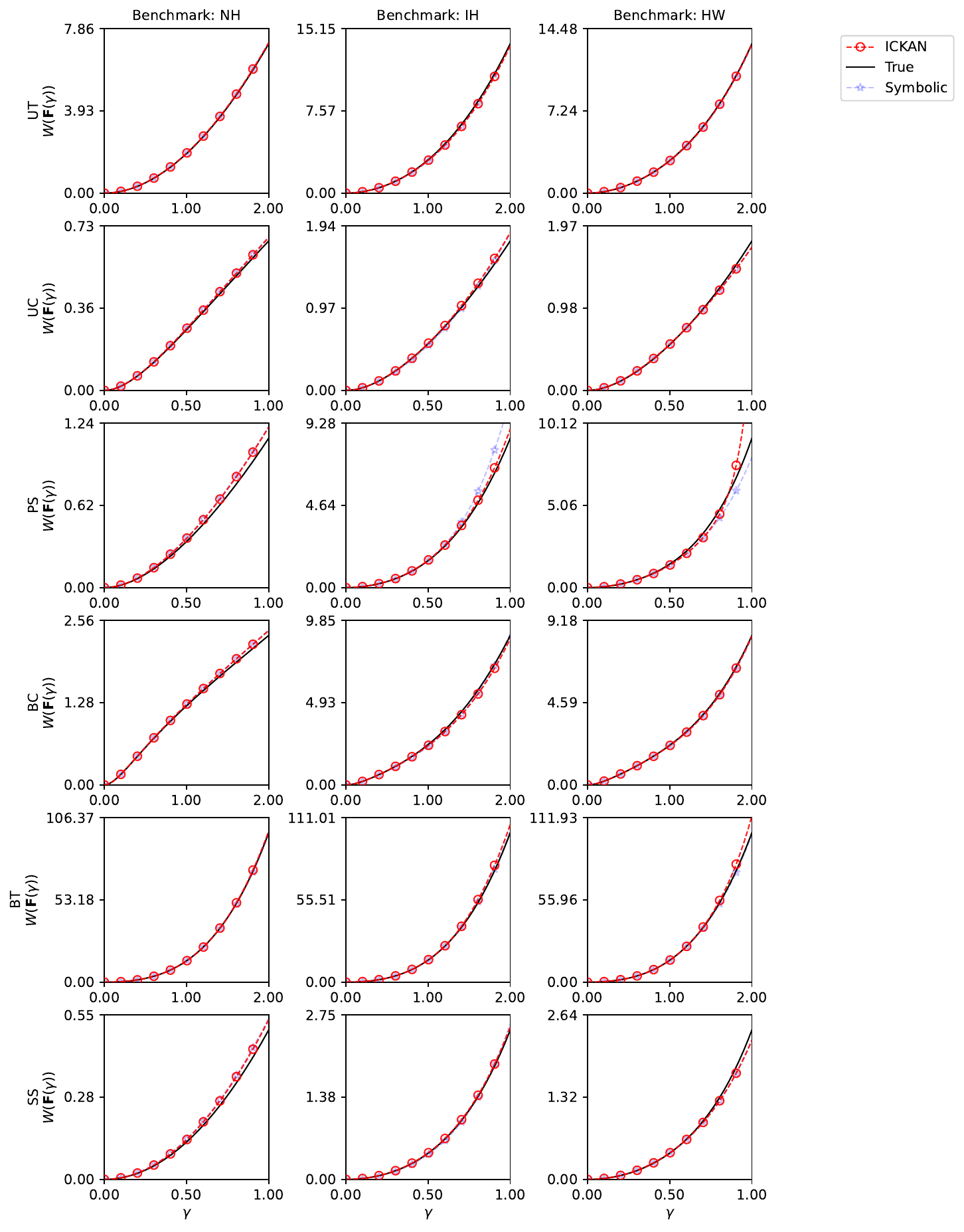}
\caption{Strain energy density $W(\bfF(\gamma))$ predictions for the deformation paths in \eqref{eq:strain_paths}, evaluated under high noise $(\sigma_u = 10^{-3})$. Results are shown for the best ICKAN-based constitutive model in the ensemble, the model obtained through symbolic regression of the ICKAN, as well as for the hidden true model, for the NH \eqref{eq:NH2}, IH \eqref{eq:IH}, and HW \eqref{eq:HW} benchmarks.}
\label{fig:NH2_IH_HW_noise=high_W}
\end{figure}

\begin{figure}
\centering
\text{Benchmark: Strain energy density predictions, high noise ($\sigma_u=10^{-3}$)}
\includegraphics[width=0.9\textwidth]{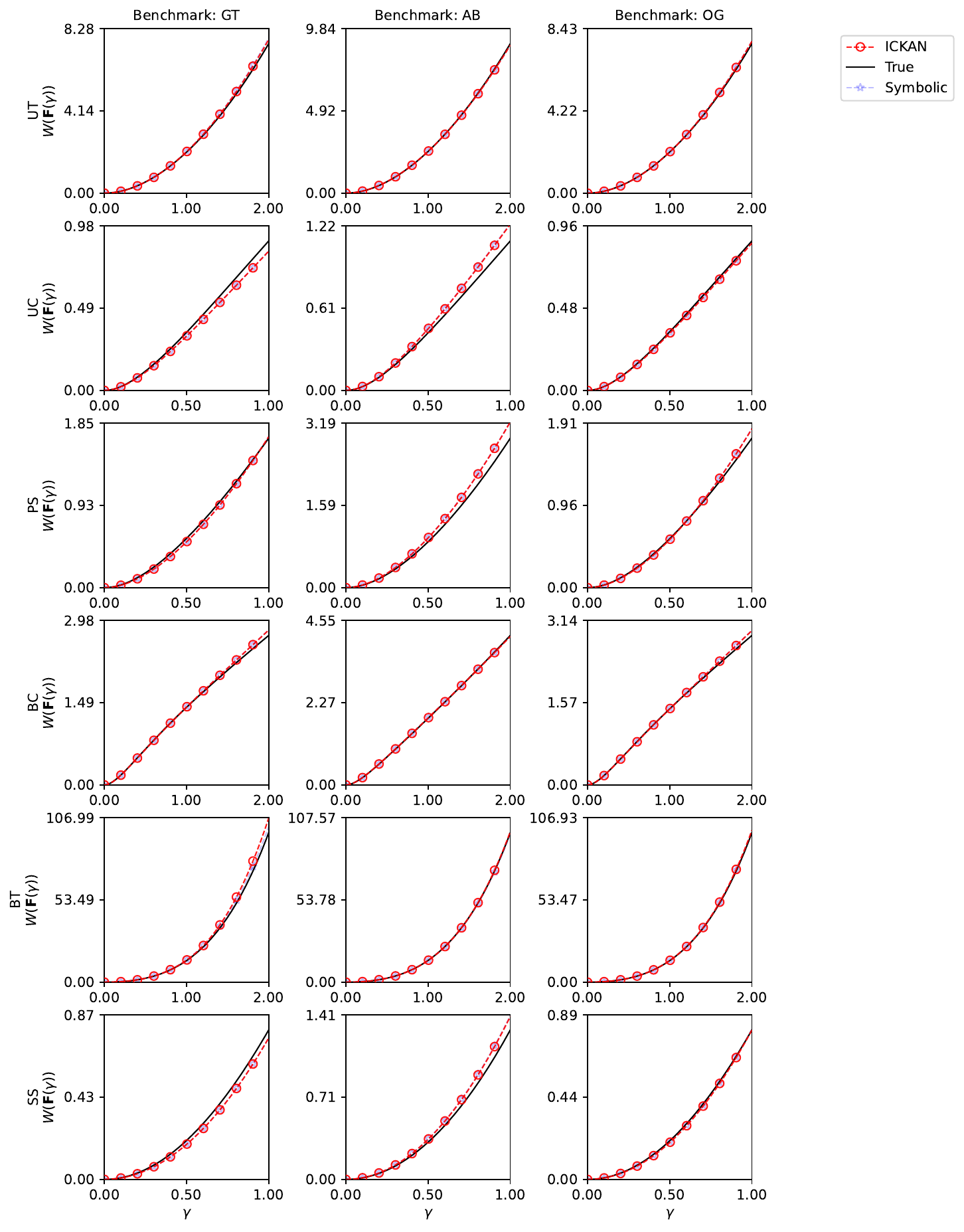}
\caption{Strain energy density $W(\bfF(\gamma))$ predictions for the deformation paths in \eqref{eq:strain_paths}, evaluated under high noise $(\sigma_u = 10^{-3})$. Results are shown for the best ICKAN-based constitutive model in the ensemble, the model obtained through symbolic regression of the ICKAN, as well as for the hidden true model, for the GT \eqref{eq:GT}, AB \eqref{eq:AB}, and OG \eqref{eq:feature_OG} benchmarks.}
\label{fig:GT_AB_OG_noise=high_W}
\end{figure}

\begin{figure}
\centering
\text{Benchmark: First Piola-Kirchhoff predictions, low noise ($\sigma_u=10^{-4}$)}
\includegraphics[width=0.9\textwidth]{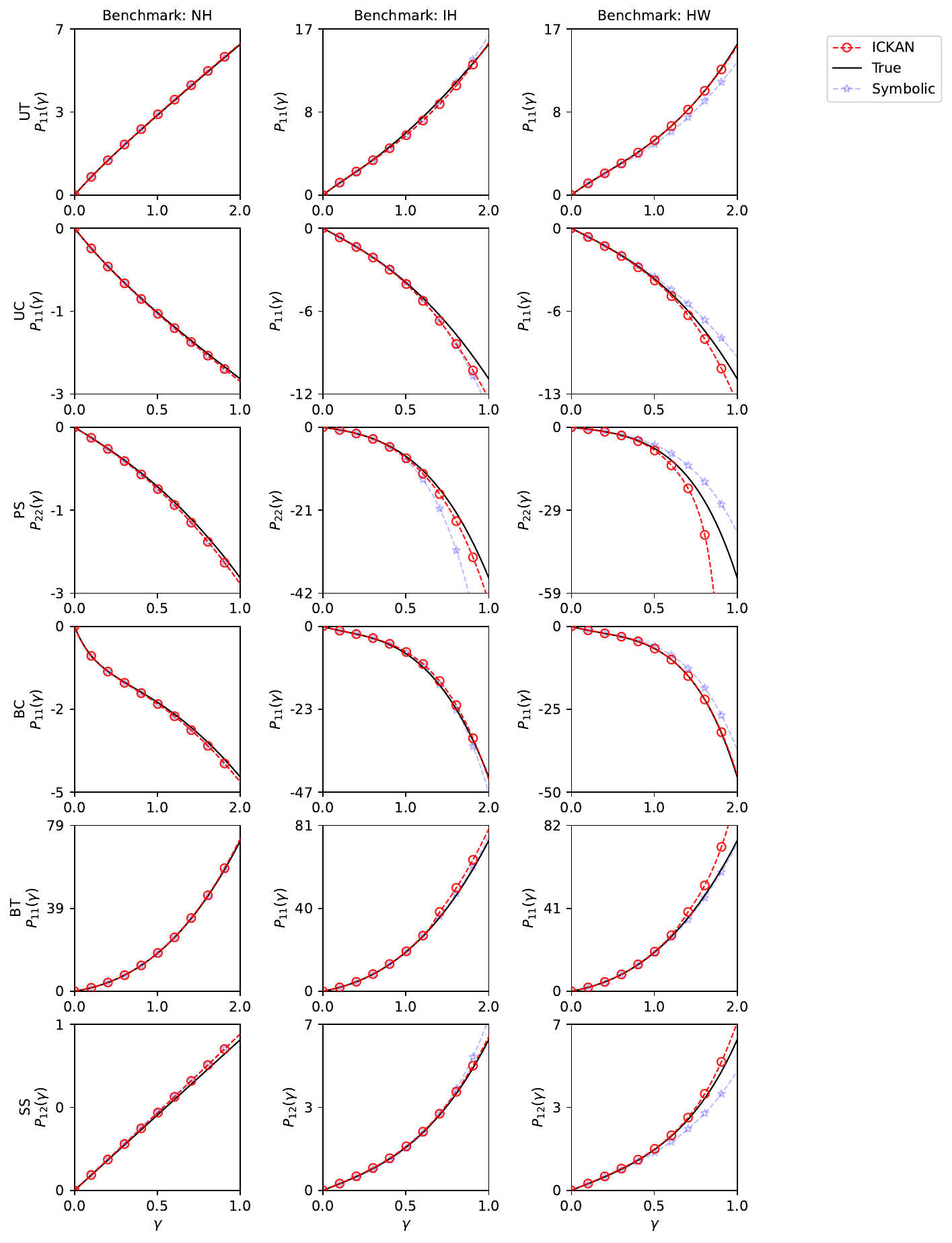}
\caption{Predicted first Piola-Kirchhoff stress $ \bfP(\bfF(\gamma)) $ components along the deformation paths specified in \eqref{eq:strain_paths} for the low noise case $(\sigma_u = 10^{-4})$. The best ICKAN-based constitutive model is shown for the NH \eqref{eq:NH2}, IH \eqref{eq:IH}, and HW \eqref{eq:HW} benchmarks, the model obtained through symbolic regression of the ICKAN, alongside the response of the true (hidden) model for comparison.}
\label{fig:NH2_IH_HW_noise=low_Pij}
\end{figure}

\begin{figure}
\centering
\text{Benchmark: First Piola-Kirchhoff predictions, low noise ($\sigma_u=10^{-4}$)}
\includegraphics[width=0.9\textwidth]{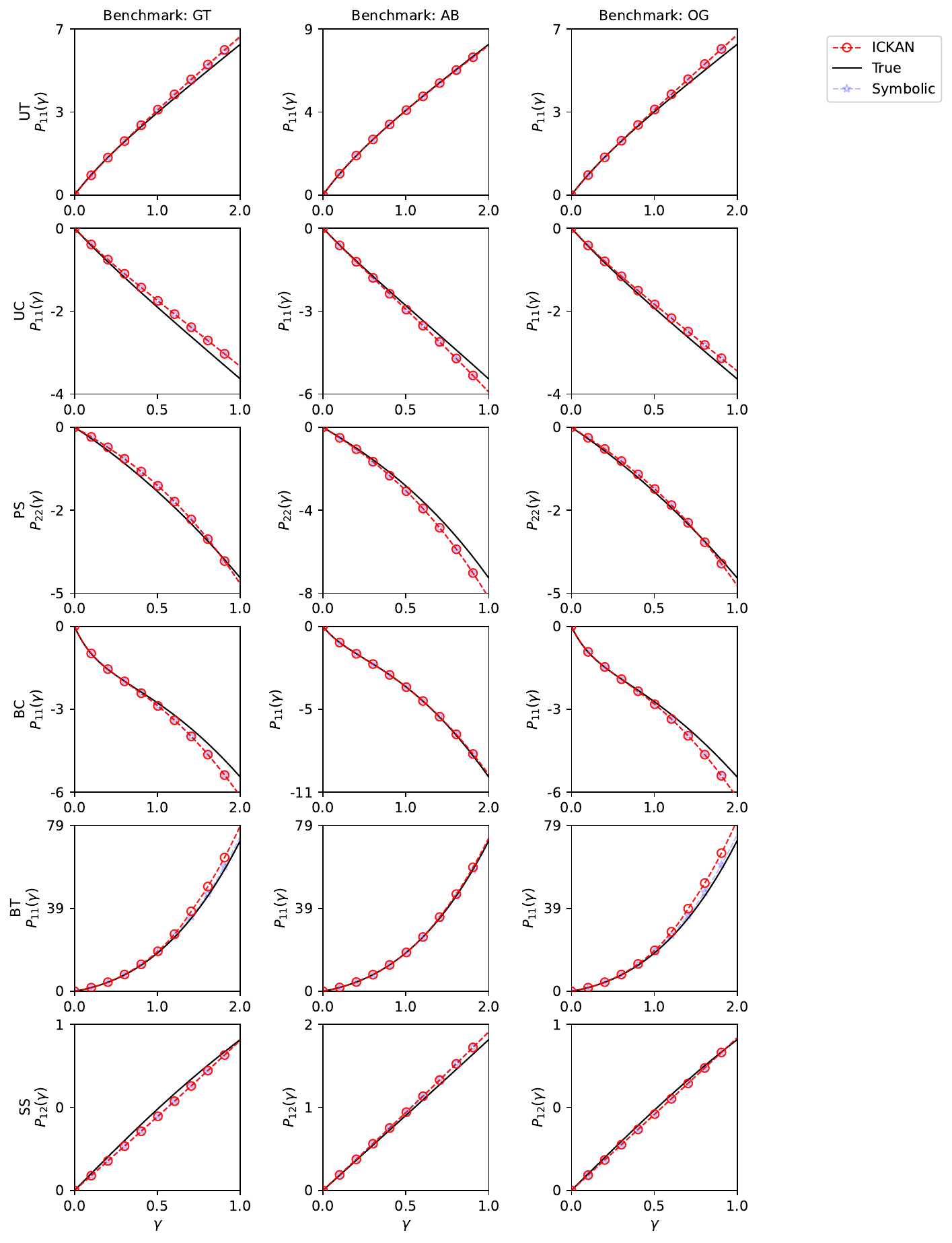}
\caption{Predicted first Piola-Kirchhoff stress $ \bfP(\bfF(\gamma)) $ components along the deformation paths specified in \eqref{eq:strain_paths} for the low noise case $(\sigma_u = 10^{-4})$. The best ICKAN-based constitutive model is shown for the GT \eqref{eq:GT}, AB \eqref{eq:AB}, and OG \eqref{eq:feature_OG} benchmarks, the model obtained through symbolic regression of the ICKAN, alongside the response of the true (hidden) model for comparison.}
\label{fig:GT_AB_OG_noise=low_Pij}
\end{figure}

\begin{figure}
\centering
\text{Benchmark: First Piola-Kirchhoff predictions, high noise ($\sigma_u=10^{-3}$)}
\includegraphics[width=0.9\textwidth]{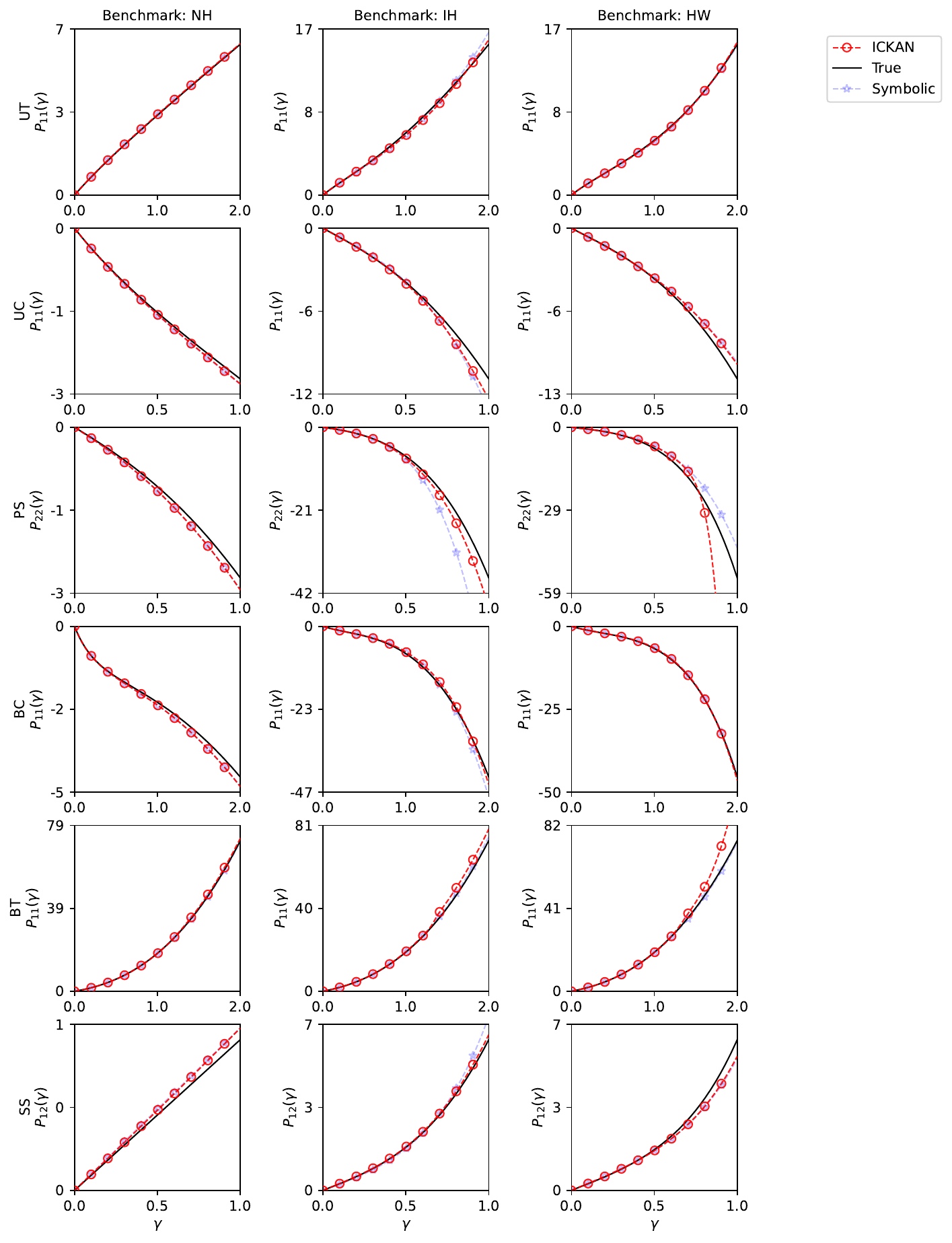}
\caption{Predicted first Piola-Kirchhoff stress $ \bfP(\bfF(\gamma)) $ components along the deformation paths specified in \eqref{eq:strain_paths} for the high noise case $(\sigma_u = 10^{-3})$. The best ICKAN-based constitutive model is shown for the NH \eqref{eq:NH2}, IH \eqref{eq:IH}, and HW \eqref{eq:HW} benchmarks, the model obtained through symbolic regression of the ICKAN, alongside the response of the true (hidden) model for comparison.}
\label{fig:NH2_IH_HW_noise=high_Pij}
\end{figure}

\begin{figure}
\centering
\text{Benchmark: First Piola-Kirchhoff predictions, high noise ($\sigma_u=10^{-3}$)}
\includegraphics[width=0.9\textwidth]{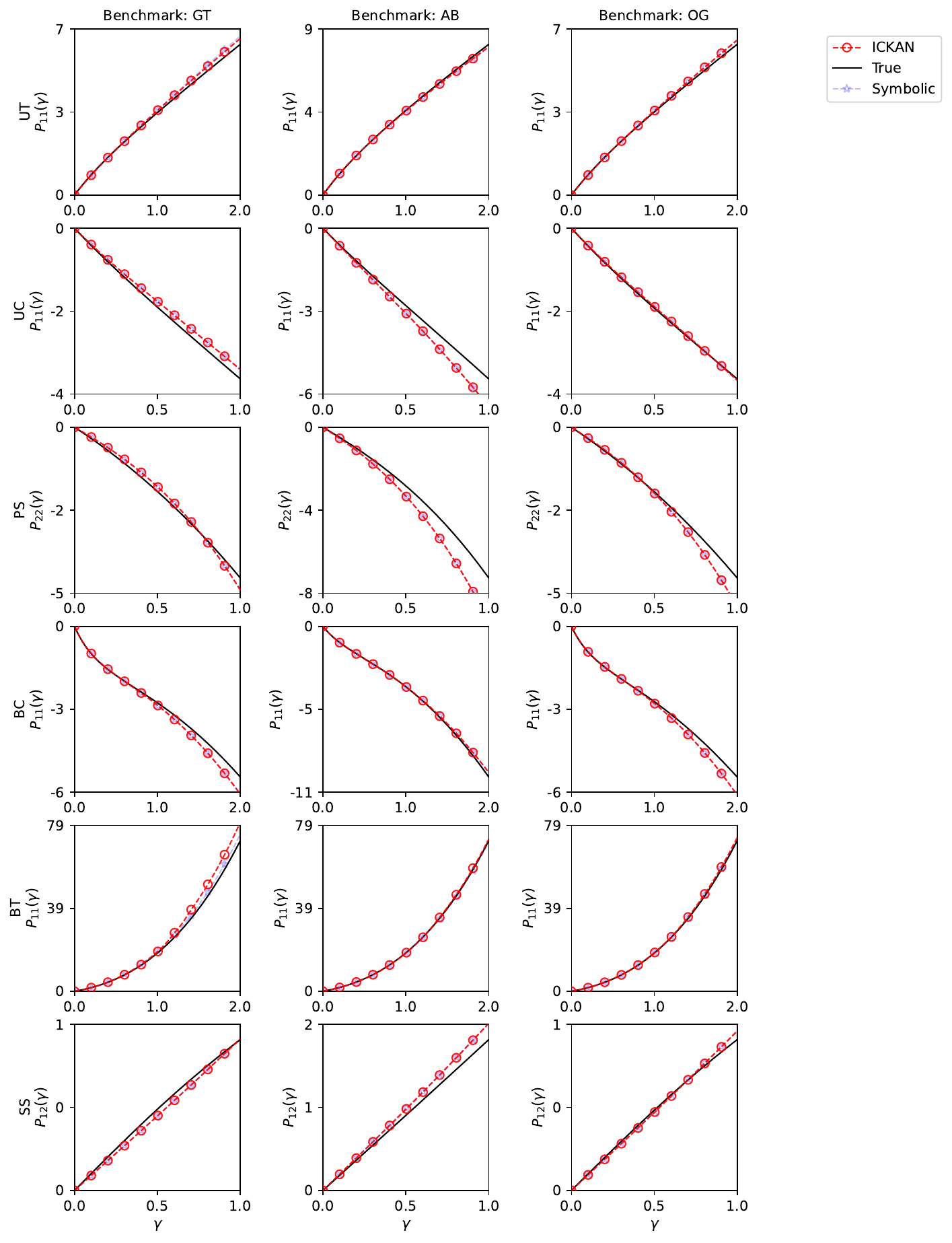}
\caption{Predicted first Piola-Kirchhoff stress $ \bfP(\bfF(\gamma)) $ components along the deformation paths specified in \eqref{eq:strain_paths} for the high noise case $(\sigma_u = 10^{-3})$. The best ICKAN-based constitutive model is shown for the GT \eqref{eq:GT}, AB \eqref{eq:AB}, and OG \eqref{eq:feature_OG} benchmarks, the model obtained through symbolic regression of the ICKAN, alongside the response of the true (hidden) model for comparison.}
\label{fig:GT_AB_OG_noise=high_Pij}
\end{figure}

\tablename~\ref{tab:strain_energy} presents the symbolic expressions of the strain energy densities obtained from the ICKANs for both low- and high-noise cases. We overlay the predictions of the symbolic models in each plot of Figures~\ref{fig:NH2_IH_HW_noise=low_W}-\ref{fig:GT_AB_OG_noise=high_Pij} for comparison with the true and ICKAN-based strain energy density and stress. \RV{In general, the symbolic regression models closely match the ICKAN predictions, and in some cases, they even provide a slightly better fit at higher strains.}  {Implementation details of the \RV{monotonic} input-convex symbolic regression are provided in \ref{sec:symbolic_regression_details}}. 

\textit{Limitations of symbolic interpretation:} Expectedly, \RV{in a few cases,} the symbolic models show a deterioration in the prediction of strain energy densities and stresses relative to the trained ICKAN models. This can be attributed to the limited expressivity of the chosen function library for symbolic regression and the compounding errors from fitting a symbolic expression to already fitted splines of the ICKAN.

\begin{table}[t]
    \centering
    \renewcommand{\arraystretch}{1.2}
    \setlength{\tabcolsep}{6pt} 
    \begin{tabular}{l l l}
        \hline
        \textbf{Model} & \textbf{Noise} & \textbf{Strain energy density ($W$)} \\
        \hline
        \rowcolor{white} \textbf{NH2 \eqref{eq:NH2}} & Truth & $0.5 (\Tilde{I}_1-3) + 1.5 (J-1)^2$ \\
        \rowcolor{green!30}& $\sigma = 10^{-4}$ & $0.5201K_1 + 1.4969K_3$ \\
        \rowcolor{blue!20} & $\sigma = 10^{-3}$ & $0.5395K_1 + 1.4986K_3$ \\
        \hline
        \rowcolor{white}\textbf{IH \eqref{eq:IH}} & Truth & $0.5 (\Tilde{I}_1-3) +  (\Tilde{I}_2-3) +  (\Tilde{I}_1-3)^2 + 1.5 (J-1)^2$ \\
        \rowcolor{green!30}& {$\sigma = 10^{-4}$} & \RV{$0.35 K_1 + 1.49 K_3  + 0.004\log(3568\exp(4.23 K_2) + 1)^2 + 0.004\log(22.43\exp(3.3 K_1) + 1)^3$} \\
        \rowcolor{blue!20}& {$\sigma = 10^{-3}$} & \RV{$0.31 K_1 + 1.48 K_3 + 0.004\log(3568\exp(4.30 K_2) + 1)^2 + 0.009\log(10.16\exp(2.5 K_1) + 1)^3$} \\
        \hline
        \rowcolor{white}\textbf{HW \eqref{eq:HW}} & Truth & $0.5 (\Tilde{I}_1-3) +  (\Tilde{I}_2-3) + 0.7 (\Tilde{I}_1-3) (\Tilde{I}_2-3) + 0.2 (\Tilde{I}_1-3)^3 + 1.5 (J-1)^2$ \\
        \rowcolor{green!30}& {$\sigma = 10^{-4}$} & \RV{$0.43K_2 + 1.48K_3 + 0.69\log(0.57\exp(1.04K_1) + 1)^2 + 0.008\log(6.64\exp(1.90K_1) + 1)^3$} \\
        \rowcolor{blue!20}& {$\sigma = 10^{-3}$} & \RV{$0.42K_2 + 1.49K_3 + 0.88\log(0.57\exp(1.04K_1) + 1)^2 + 0.591\log(0.57\exp(1.04K_1) + 1)^2$} \\
        \hline
        \rowcolor{white}\textbf{GT \eqref{eq:GT}} & Truth & $0.5 (\Tilde{I}_1-3) + \log(\Tilde{I}_2/3) + 1.5 (J-1)^2$ \\
        \rowcolor{green!30}& $\sigma = 10^{-4}$ & $0.65K_1 + 0.03K_2 + 0.44K_3 + 1.58\log(35.20\exp(0.69K_3) + 1)$ \\
        \rowcolor{blue!20}& $\sigma = 10^{-3}$ & $0.62K_1 + 0.04K_2 + 0.78K_3 + 1.09\log(16.45\exp(0.72K_3) + 1)$ \\
        \hline
        \rowcolor{white}\textbf{AB \eqref{eq:AB}} & Truth & Refer to \eqref{eq:AB} \\
        \rowcolor{green!30}& $\sigma = 10^{-4}$ & $1.1959K_1 +0.0533K_2 +1.4905K_3$ \\
        \rowcolor{blue!20}& $\sigma = 10^{-3}$ & $1.0861K_1 + 0.1140K_2 + 1.4790K_3$ \\
        \hline
        \rowcolor{white}\textbf{OG \eqref{eq:feature_OG}} & Truth &  Refer to \eqref{eq:feature_OG} \\
        \rowcolor{green!30}& $\sigma = 10^{-4}$ & $0.76K_1 + 0.001K_2 + 0.84K_3 + 1.09\log(13.46\exp(0.66K_3)+1)$ \\
        \rowcolor{blue!20}& $\sigma = 10^{-3}$ & $0.69K_1 + 0.042K_2 + 1.35K_3 + 0.23\log(17.64\exp(0.73K_3)+1)$ \\
        \hline
    \end{tabular}
    \caption{Strain energy density of the (true) hidden and discovered symbolic expressions by the ICKAN-based material models for different noise levels $\sigma_u$. Note that the symbolic expressions do not satisfy $W(\bfF=\bfI)=0$; this can be manually fixed by re-introducing the energy correction term $W^0$ (see \eqref{zero-energy}). However, they do satisfy the condition $P(\bfF=\bfI)=0$ identically (see \eqref{zero-stress}).}
    \label{tab:strain_energy}
\end{table}

\section{Conclusion}
{We introduced \RV{monotonic} Input-Convex Kolmogorov–Arnold Networks (ICKANs) for learning hyperelastic constitutive laws. By integrating convexity \RV{and monotonicity} constraints directly into the KAN architecture, our approach guarantees that the predicted strain energy density satisfies polyconvexity. At the same time, the use of trainable spline-based activation functions allows for a compact and interpretable representation of the constitutive model. We further demonstrated that ICKANs can be trained in an unsupervised manner on realistically measurable data in the form of strain fields and global reaction forces.

Extensive numerical experiments demonstrated that the ICKAN-based models accurately capture the nonlinear stress-strain behavior of hyperelastic materials, even under varying levels of noise in the data. The predicted strain energy densities and first Piola–Kirchhoff stresses closely match the ground truth for numerous material models across multiple deformation paths. Furthermore, finite element simulations on a validation specimen with a complex geometry confirmed the generalization capability of our method, as the ICKAN-based constitutive models produced robust and physically consistent results.

Our findings indicate that the ICKAN framework successfully overcomes key limitations of ICNN-based models by combining physical consistency with enhanced interpretability and efficiency. Future work will explore extending the framework to model additional material behaviors such as viscoelasticity and plasticity, as well as integrating uncertainty quantification to further bolster model reliability in real-world applications.} \RV{Additionally, the ICKAN approach can be applied to any machine learning task that ICNNs, as introduced by \citet{amos_input_2017}, can address, thereby benefiting the broader machine learning community.}

\section*{Declaration of competing interest}
The authors declare that they have no known competing financial interests or personal relationships that could have appeared to influence the work reported in this paper.

\section*{Acknowledgement}
Y. Guo sincerely thanks Dr. Ye Ji and Chenghuai Lin (MSc.) for their insightful discussions on convex B-splines.

\section*{Data and code availability}
The data and codes generated in the current study are freely available at \quad \url{https://github.com/mmc-group/ICKANs/}.

\bibliographystyle{elsarticle-harv}
\bibliography{Bib}

\clearpage

\appendix

\section{Details on enforcing convexity and monotonicity}

\subsection{Proof of the non-decreasing convexity condition}\label{sec:convexity proof}

The derivative of a spline basis function can be computed as {\citep{butterfield1976computation}}:
\begin{equation}
\frac{d B_{i,k}(x)}{dx} = \frac{k}{t_{i+k}-t_i} B_{i,k-1}(x) -\frac{k}{t_{i+k+1}-t_{i+1}}  B_{i+1,k-1}(x).
\end{equation}
This implies that
\begin{equation}
\begin{split}
\frac{d \psi(x)}{dx} &= \sum_{i=1}^{n_b} c_i \left( \frac{k}{t_{i+k}-t_i} B_{i,k-1}(x)-\frac{k}{t_{i+k+1}-t_{i+1}} B_{i+1,k-1}(x) \right) \\
&=k \sum_{i=0}^{n_b-1} \frac{c_{i+1}}{t_{i+k+1}-t_{i+1}} B_{i+1,k-1}(x)-k \sum_{i=1}^{n_b} \frac{c_i}{t_{i+k+1}-t_{i+1}} B_{i+1,k-1}(x) \\
&=\frac{k c_1}{t_{k+1}-t_1} B_{1,k-1}(x) + k \sum_{i=1}^{n_b-1} \frac{c_{i+1}-c_i}{t_{i+k+1}-t_{i+1}} B_{i+1,k-1}(x)-\frac{k c_{n_b}}{t_{n_b+k+1}-t_{n_b+1}} B_{n_b+1,k-1}(x)
\end{split}
\end{equation}
Note that by definition of the basis functions, the first and last terms are zero. For uniform B-splines, the knots are equally spaced by interval $s = t_{i+1}-t_i > 0$, for $i \in [1,n_b-1]$. Hence the first order derivative can be simplified as
\begin{equation}
\frac{d \psi(x)}{dx}=\frac{1}{s} \sum_{i=1}^{n_b-1} \left( c_{i+1}-c_i \right) B_{i+1,k-1}(x).
\end{equation}
Since basis functions are by definition non-negative, the non-decreasing requirement
\begin{equation}
\frac{d \psi(x)}{dx} \geq 0
\end{equation}
leads to 
\begin{equation}
c_{i+1}-c_i \geq 0, \qquad \forall i \in [1,n_b-1].
\end{equation}

The second-order derivative of the basis function is:
\begin{equation}
\begin{split}
\frac{d^2 B_{i,k}(x)}{dx^2} &= \frac{k}{t_{i+k} - t_i} \left(\frac{k-1}{t_{i+k-1} - t_i} B_{i,k-2}(x) - \frac{k-1}{t_{i+k} - t_{i+1}} B_{i+1,k-2}(x) \right) - \\
& \frac{k}{t_{i+k+1} - t_{i+1}} \left(\frac{k-1}{t_{i+k} 
- t_{i+1}} B_{i+1,k-2}(x) - \frac{k-1}{t_{i+k+1} - t_{i+2}} B_{i+2,k-2}(x) \right) \\
&=\frac{1}{s^2} \left( B_{i,k-2}(x) -2 B_{i+1,k-2}(x) + B_{i+2,k-2}(x)\right). \\
\end{split}
\end{equation}
Substituting this into $ \psi(x) $ and considering the definition of the basis functions, we have:
\begin{equation}
\begin{split}
\frac{d^2 \psi(x)}{dx^2} &= \sum_{i=1}^{n_b} c_i \frac{d^2 B_{i,k}(x)}{dx^2} \\
&= \frac{1}{s^2} \left( \sum_{i=-1}^{n_b-2} c_{i+2} B_{i+2,k-2}(x) -2 \sum_{i=0}^{n_b-1} c_{i+1} B_{i+2,k-2}(x) + \sum_{i=1}^{n_b} c_{i} B_{i+2,k-2}(x) \right) \\
&= \frac{1}{s^2} \left( \sum_{i=1}^{n_b-1} (c_{i+2} -2 c_{i+1} + c_{i}) B_{i+2,k-2}(x) \right).
\end{split}
\end{equation}

Convexity condition requires
\begin{equation}
\frac{d^2 \psi(x)}{dx^2} \geq 0.
\end{equation}
Since $ B_{i+2,k-2}(x) \geq 0 $, convexity is ensured if:

\begin{equation}
c_{i+2} - 2c_{i+1} +c_i  \geq 0, \quad \forall i \in [1, n_b-2].
\end{equation}

\subsection{Implementation of convexity and monotonicity constraints}\label{sec:constraints}
Let $\mathbf{p}=\{p_i\}_{i=1}^{n_b}$ be a sequence of randomly initialized coefficients, A sequence of convex coefficients $\mathbf{c} = \{c_i\}_{i=1}^{n_b}$ can be derived from $\mathbf{p}$ through the following steps:
\begin{equation}
\begin{split}
\mathbf{h} &= \{h_i\}_{i=1}^{n_b}, \quad \text{with} \quad
    h_i=
        \begin{cases}
        p_1, & \text{if } i=1, \\
        \text{max}(0,p_i), & \text{otherwise},
        \end{cases} \\
\mathbf{d} &= \{d_i\}_{i=1}^{n_b}, \quad \text{with} \quad
    d_i=
        \begin{cases}
        h_1, & \text{if } i=1, \\
        \sum_{{i=2}}^i h_j, & \text{otherwise},
        \end{cases} \\
\mathbf{c} &= \{c_i\}_{i=1}^{n_b}, \quad \text{with} \quad
    c_i=\sum_{j=1}^i d_j.
\end{split}
\end{equation}
Since $\mathbf{d} = \{d_i\}_{{i=2}}^{n_b}$ is the cumulative sum of a sequence of non-negative values $\mathbf{h} = \{h_i\}_{{i=2}}^{n_b}$, it follows that:

\begin{equation}
c_{i+1} - c_i = d_{i+1} \geq d_i = c_i - c_{i-1} \geq 0, \quad \forall i \in [2, n_b-1].
\end{equation}
Thus, the sequence $\mathbf{c} = \{c_i\}_{i=1}^{n_b}$ is non-decreasing and convex.  

This algorithm enforces a hard constraint on the coefficients, ensuring convexity by construction. Consequently, backpropagation can be performed efficiently during training without introducing additional computational overhead.

\subsection{Linear Extrapolation at the endpoints} \label{sec:extrapolation}
As illustrated in \figurename~\ref{fig:spline}b, convexity cannot be guaranteed outside the B-spline domain  $[t_{k+1}, t_{m_b-k}]$. To smoothly extend the curve $\psi(x)$ beyond these boundaries, we apply a linear extrapolation that preserves the local behavior at the endpoints.

For the left endpoint at $ x = t_{k+1} $, choose two sample points using a small offset $1 \gg \varepsilon > 0$:
\be
{x_0 = t_{k+1}}, \quad x_1 = t_{k+1} + \varepsilon.
\ee
Evaluating the curve at these points gives:
\be
y_0 = \psi(x_0),\quad y_1 = \psi(x_1).
\ee
We fit a linear function of the form:
\be
\psi_{\text{left}}(x) = a_{\text{left}} (x-t_{k+1})+b_{\text{left}}.
\ee
Using the finite difference approximation, the coefficients are determined as:
\be
{a_{\text{left}} = \frac{y_1 - y_0}{\varepsilon}= \frac{\psi(t_{k+1}+\varepsilon) - \psi(t_{k+1})}{\varepsilon}, \qquad b_{\text{left}} = \psi(t_{k+1}).}
\ee
Thus, for $ x < t_{k+1} $, the extrapolated curve is given by
\be
{\psi_{\text{left}}(x) =  \frac{\psi(t_{k+1}+\varepsilon) - \psi(t_{k+1})}{\varepsilon} (x-t_{k+1}) + \psi(t_{k+1}).}
\ee

Similarly, for the right endpoint at $ x = t_{m_b-k} $, extrapolation function can be evaluated as:
\be
{\psi_{\text{right}}(x) =  \frac{\psi(t_{m_b-k}) - \psi(t_{m_b-k} - \varepsilon)}{\varepsilon} (x-t_{m_b-k}) + \psi(t_{m_b-k}), \qquad x > t_{m_b-k}.}
\ee

By applying the above linear extrapolation at both endpoints, the extended B-spline curve $ \bar{\psi}(x) $ is defined as
\be
{
\bar{\psi}(x) =
\begin{cases}
\frac{\psi(t_{k+1}+\varepsilon) - \psi(t_{k+1})}{\varepsilon} (x-t_{k+1}) + \psi(t_{k+1}), & x < t_{k+1}, \\
\psi(x), & t_{k+1} \le x \le t_{m_b-k}, \\
\frac{\psi(t_{m_b-k}) - \psi(t_{m_b-k} - \varepsilon)}{\varepsilon} (x-t_{m_b-k}) + \psi(t_{m_b-k}), & x > t_{m_b-k}.
\end{cases}}
\ee
This approach ensures a smooth extension of the B-spline, preserving both function continuity ($C^0$) and first-order smoothness ($C^1$) at the boundaries, thereby maintaining local shape consistency. \RV{Note that while we can ensure physical consistency for $W$ outside the spline domain using linear extrapolation, it may not be the most ideal solution and could hinder generalization.}

\section{ICKAN spline grid initialization} \label{sec:grid_init}

In our ICKAN implementation, the  natural definition domain bounds of the splines are set  during the initialization phase to ensure they cover the expected range of  input values.

This process begins by constructing  dummy inputs in the form of $\bfz^{(0)}$---with each dimension's values set in the sufficiently large range of $[-5,25]$ discretized along 100 uniformly spaced points. These dummy inputs are passed through the first KAN layer ($r=1$) and the outputs, i.e., $\bfz^{(1)}$, are recorded. The range of values for $z^{(1)}_j$ (with $j\in\{1,\dots,n_0\}$) is computed, which is then set as the natural definition domain for $\phi_{0,i,j}$ for all $i\in\{1,\dots,n_1\}$.  

Similar to the previous step, the new range of values for each dimension of  $\bfz^{(1)}$ is set to the respective natural definition domains and new dummy inputs  are created for the second KAN layer ($r=2$). The process is then subsequently repeated for the following layers in a similar fashion.

Once this initialization step is complete for all KAN layers, the bounds remains fixed and are not updated during training or inference. 

\section{Training details}\label{sec:training-details}
The Algorithm \ref{alg:pseudocode} (adapted from NN-EUCLID \citep{thakolkaran2022nn}) summarizes the unsupervised training of the ICKAN-based constitutive models. All parameters used for data generation and training are found in \tablename~\ref{tab:parameters}.

\begin{algorithm} 
\caption{Unsupervised training of the ICKAN-based constitutive models}
\label{alg:pseudocode}
\begin{algorithmic}[1]
\State \textbf{Input:} Point-wise displacement data $\calU = \{\bfu^{a,t} \in \mathbb{R}^2 : a = 1, \dots, n_n; t=1,\dots,n_t\}$ 
\State \textbf{Input:} Global reaction forces $\{R^{\beta,t}: \beta=1,\dots,n_\beta; t=1,\dots,n_t\}$
\State \text{Randomly initialize ICKAN parameters: $\calQ$.}
\State \text{Initialize learning rate scheduler}
\State \text{Initialize Adam optimizer with parameters $\calQ$ and learning rate scheduler}
    \For{$e = 1,\dots,n_e$} \Comment{Training epochs}
        \State Loss: $\ell\gets 0$ \Comment{Initialize loss for current epoch}
        \For{$t = 1,\dots,n_t$} \Comment{Iterate over snapshots}
            \For{each element in mesh}
                \State $W^{0} \gets W^{\text{ICKAN}}_{\calQ}\vert_{\bfF=\bfI}$ \Comment{see \eqref{zero-energy}}
                \State $W \gets W^{\text{ICKAN}}_{\calQ} +  W^{0} $\Comment{see \eqref{eq:ansatz}}
                \State  $\boldsymbol{P} \gets \frac{\partial  W}{\partial \boldsymbol{F}}$\Comment{see \eqref{eq:PK_ansatz}}
                
            \EndFor
            \For{$a = 1,\dots,n_a$}
                \For{$i = 1,2$}
	                \State Compute force $f^{a,t}_i$ using \eqref{reducedWeakForm}
                \EndFor
            \EndFor
            \For{$(a,i) \in  \calD^\text{free}$}
                \State $\ell\gets\ell + \left(f^{a,t}_i\right)^2 $ \Comment{force balance at free degrees of freedom; see \eqref{loss}}
            \EndFor
            \For{$\beta= 1,\dots, n_\beta$}
                \State $r^{\beta,t} \gets 0$
                \For{$(a,i) \in  \calD^\text{fix}_\beta$}
                    \State $r^{\beta,t} \gets r^{\beta,t} + f^{a,t}_i$ \Comment{see \eqref{loss}}
                \EndFor
                \State $\ell\gets \ell + \left(R^{\beta,t}-r^{\beta,t}\right)^2$ \Comment{force balance at fixed degrees of freedom; see \eqref{loss}}
            \EndFor
        \EndFor
        \State Compute gradients $\partial \ell/\partial \calQ$ using automatic differentiation
        \State Update $\calQ$ with Adam optimizer using gradients $\partial \ell/\partial \calQ$.
        \State Update learning rate with learning rate scheduler based on epoch number $e$ 
    \EndFor
    \State \textbf{Output:} Trained ICKAN model $W^\text{ICKAN}_{\calQ}$
\end{algorithmic}
\end{algorithm}

\begin{table}[t]
\centering
\caption{List of parameters and hyperparameters used for the data generation and benchmarks.} 
\label{tab:parameters}
\begin{tabular}{lcc}
\hline
\multicolumn{1}{l}{\textbf{Parameter}} & \textbf{Notation} & \textbf{Value} \\ \hline
\textit{Training specimen:}\\
$\quad$ Number of nodes in mesh for FEM-based data generation  & - & $63,601$  \\
$\quad$ Number of nodes in data available for learning  & $n_n$ & $1,441$\\
$\quad$ Number of reaction force constraints & $n_{\beta}$   & $4$   \\ 
$\quad$ Number of data snapshots for NH, GT    & $n_{t}$     & 3 \\
$\quad$ Number of data snapshots for IH, HW    & $n_{t}$     & 8 \\
$\quad$ Number of data snapshots for AB    & $n_{t}$     & 10 \\
$\quad$ Number of data snapshots for OG    & $n_{t}$     & 6 \\

$\quad$ Loading parameter for NH, GT, IH, HW  & $\delta$  & $\{0.1 \times t: t=1,\dots,n_t\}$ \\
$\quad$ Loading parameter for AB, OG  & $\delta$  & $\{0.05 \times t: t=1,\dots,n_t\}$ \\ \hline
\textit{Validation specimen:}\\
$\quad$ Number of nodes in the mesh  & - & $4,908$ \\
$\quad$ Loading parameter   & $\delta$  & $\{0.01 \times t: t=1,\dots,100\}$ \\ \hline
\textit{ICKAN hyperparameters:}\\
$\quad$ Order of B-splines & $k$    & $5$\\
$\quad$ Number of coefficients per trainable activations & $n_b$    & $17$\\
$\quad$ Number of knots & $m_b$    & $23$\\
$\quad$ Number of hidden layers & $R-1$    & $1$\\
$\quad$ Number of trainable activations in hidden layer $(R-1)$  & $n_0\times n_{R-1}$    & $3\times 2 = 6$  \\
$\quad$ Number of trainable activations in output layer $(R)$  & $n_{R-1}\times n_{R}$    & $2\times 1 = 2$  \\
$\quad$ Number of trained ICKAN models  & $n_{e}$    & $10$  \\
$\quad$ Optimizer  & $-$    & Adam  \\
$\quad$ Epochs  & $-$    & 1000  \\
$\quad$ Learning rate scheduler  & $-$    & cyclic  \\
$\quad$ Base learning rate  & $-$    & 0.001 \\
$\quad$ Maximum learning rate  & $-$    & 0.1 \\
$\quad$ Learning rate scheduler steps  & $-$    & 50 \\
$\quad$ Symbolic regression balance parameter  & $\lambda_\text{sym}$    & 0.8 
\end{tabular}
\end{table}

\section{Learning material models without convexity constraints}\label{sec:nonconvex}
\begin{figure}
\centering
\text{Benchmark: Strain energy density predictions, high noise ($\sigma_u=10^{-3}$)}
\includegraphics[width=0.9\textwidth]{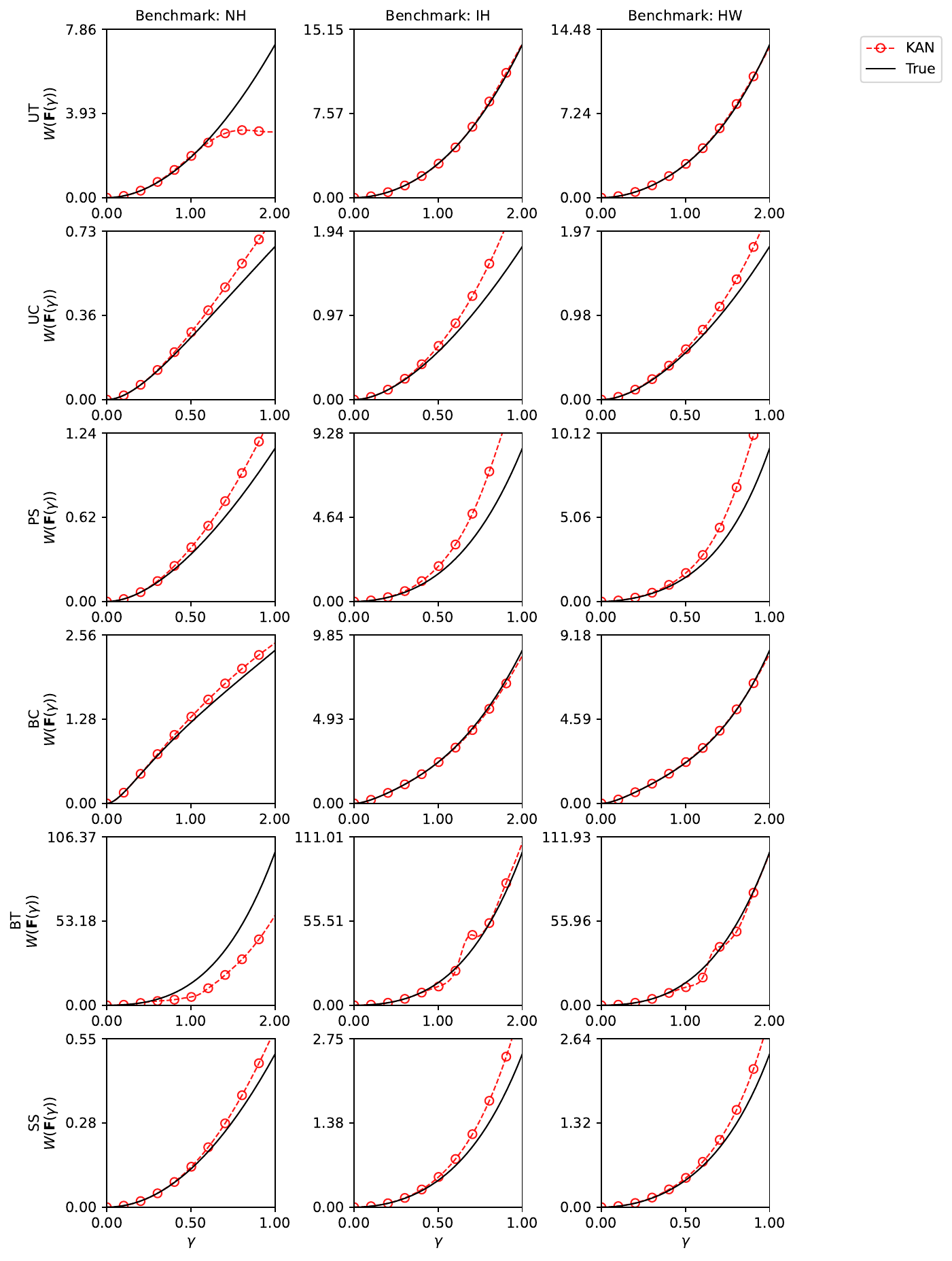}
\caption{Strain energy density $W(\bfF(\gamma))$ predictions for the deformation paths in \eqref{eq:strain_paths}, evaluated under high noise $(\sigma_u = 10^{-3})$. Results are shown for a vanilla KAN-based model, as well as for the hidden true model, for the NH \eqref{eq:NH2}, IH \eqref{eq:IH}, and HW \eqref{eq:HW} benchmarks.}
\label{fig:KAN-NH2_IH_HW_noise=high_W}
\end{figure}

\begin{figure}
\centering
\text{Benchmark: First Piola-Kirchhoff predictions, high noise ($\sigma_u=10^{-3}$)}
\includegraphics[width=0.9\textwidth]{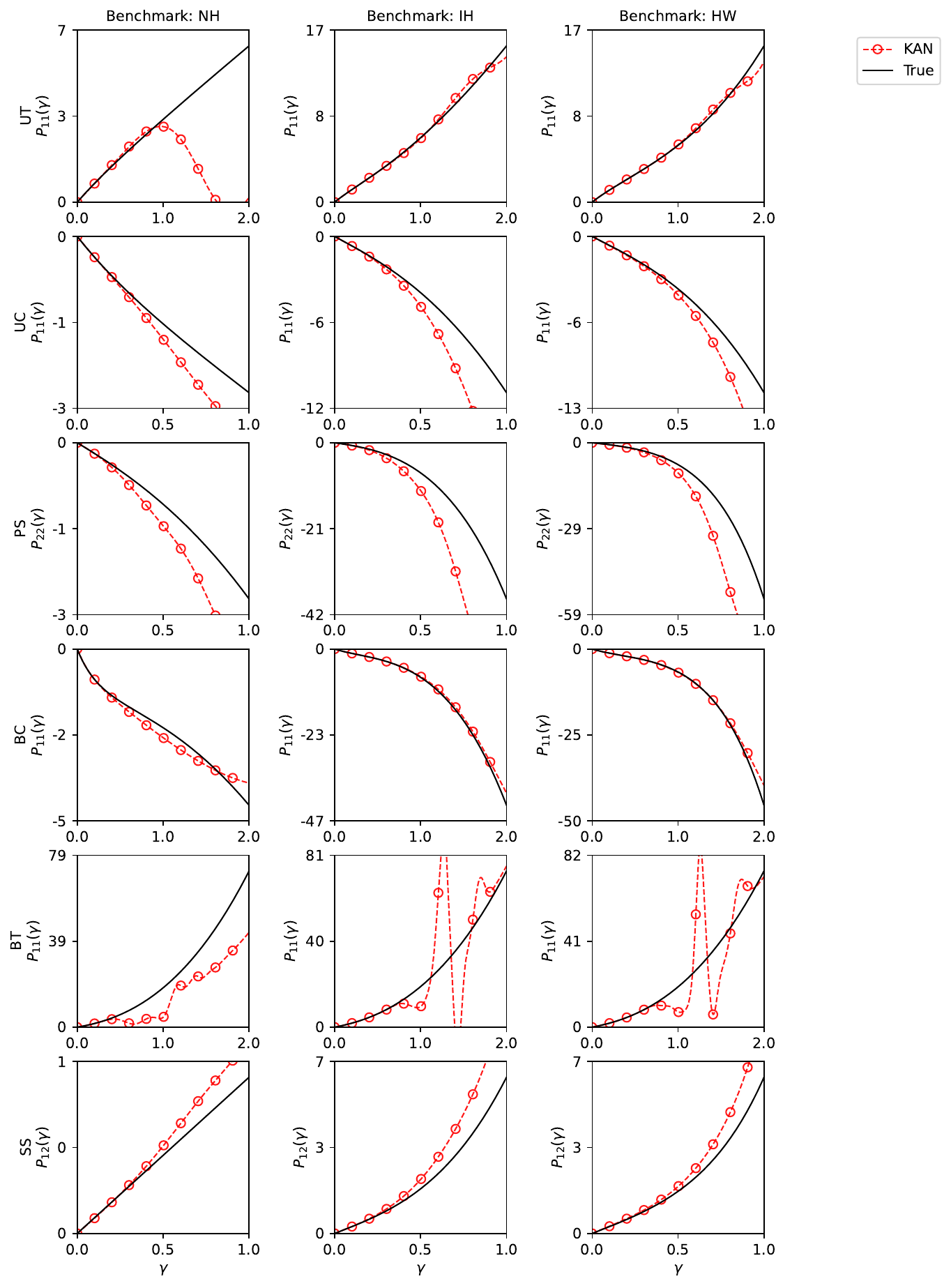}
\caption{Predicted first Piola-Kirchhoff stress $ \bfP(\bfF(\gamma)) $ components along the deformation paths specified in \eqref{eq:strain_paths} for the high noise case $(\sigma_u = 10^{-3})$. The response of a trained vanilla KAN-based model is shown for the NH \eqref{eq:NH2}, IH \eqref{eq:IH}, and HW \eqref{eq:HW} benchmarks, alongside the response of the true (hidden) model for comparison.}
\label{fig:KAN-NH2_IH_HW_noise=low_Pij}
\end{figure}

{\figurename~\ref{fig:KAN-NH2_IH_HW_noise=high_W}-\ref{fig:KAN-NH2_IH_HW_noise=low_Pij} illustrate the constitutive response for three representative benchmarks when the ICKAN model in \eqref{eq:ansatz} is replaced with a standard KAN without the convexity and monotonicity constraints while maintaining the same network architecture. We kept the initial update of the knot range (outlined in \ref{sec:grid_init}). Here, we included the bias function in the construction of the univariate function to improve the learning (as originally suggested for KANs by \cite{liu2024kan}), i.e.,
\be
\phi(x) = w_b b(x) + w_s \psi(x), \qquad \text{where }\quad b(x)=\frac{x}{1+\exp{(\RV{-x})} } \quad \text{is the SiLU activation \citep{elfwing2018sigmoid}}
\ee
and $w_b$ is a trainable scalar weight.
Note that we did not perform any other hyperparameter tuning in this case.}

\section{\RV{Comparison with EUCLID}}\label{sec:euclid}

\RV{\citet{flaschel_unsupervised_2021} have previously introduced the EUCLID framework for discovering interpretable and analytical hyperelastic models via sparse (linear) regression  on a large catalog of candidate functions in an unsupervised setting. Since the candidate functions are predefined, EUCLID inherently provides interpretable models. However, its expressivity may be limited by the chosen set of functions used for sparse regression.}

\RV{Here, we compare the vanilla EUCLID framework with ICKANs. The EUCLID implementation used in this work is directly adapted from \cite{flaschel_unsupervised_2021} (code available at: \href{https://github.com/EUCLID-code/EUCLID-hyperelasticity}{github.com/EUCLID-code/EUCLID-hyperelasticity}). In consultation with the authors of \cite{flaschel_unsupervised_2021}, we changed the penalty parameter $\lambda_p$ therein from 0.0001 to 0.0125 to adapt the implementation to our data. For details such as the methodology, hyperparameters, and the catalog of candidate functions, please refer to \cite{flaschel_unsupervised_2021}.}

\begin{figure}[ht]
\raggedright
{Benchmark: First Piola-Kirchhoff predictions, high noise ($\sigma_u=10^{-3}$)}

\vspace{0.5em}

\begin{picture}(0,0)
    \put(245,-250){
        \begin{minipage}[t]{0.5\textwidth}
        \small
        \textbf{{Discovered model formulations via EUCLID:}} \\[0.5em]
        \be
        \begin{split}
        W_\text{IH} = & 1.6197 (\Tilde{I}_2-3) + 1.5260 (\Tilde{I}_1-3)(\Tilde{I}_2-3) + 0.0193(\Tilde{I}_1-3)^4\\\notag
         &+ 1.4494 (J-1)^2 \\\notag
        W_\text{AB} = & 1.3677 (\Tilde{I}_1-3) + 0.0432 (\Tilde{I}_2-3)^2 + 0.0120(\Tilde{I}_1-3)^2(\Tilde{I}_2-3) \\ \notag
        &+ 1.5120 (J-1)^2 - 0.3790 \log(\Tilde{I}_2/3)\notag
        \end{split}
        \ee
        \end{minipage}
    }
\end{picture}

\includegraphics[width=0.68\textwidth]{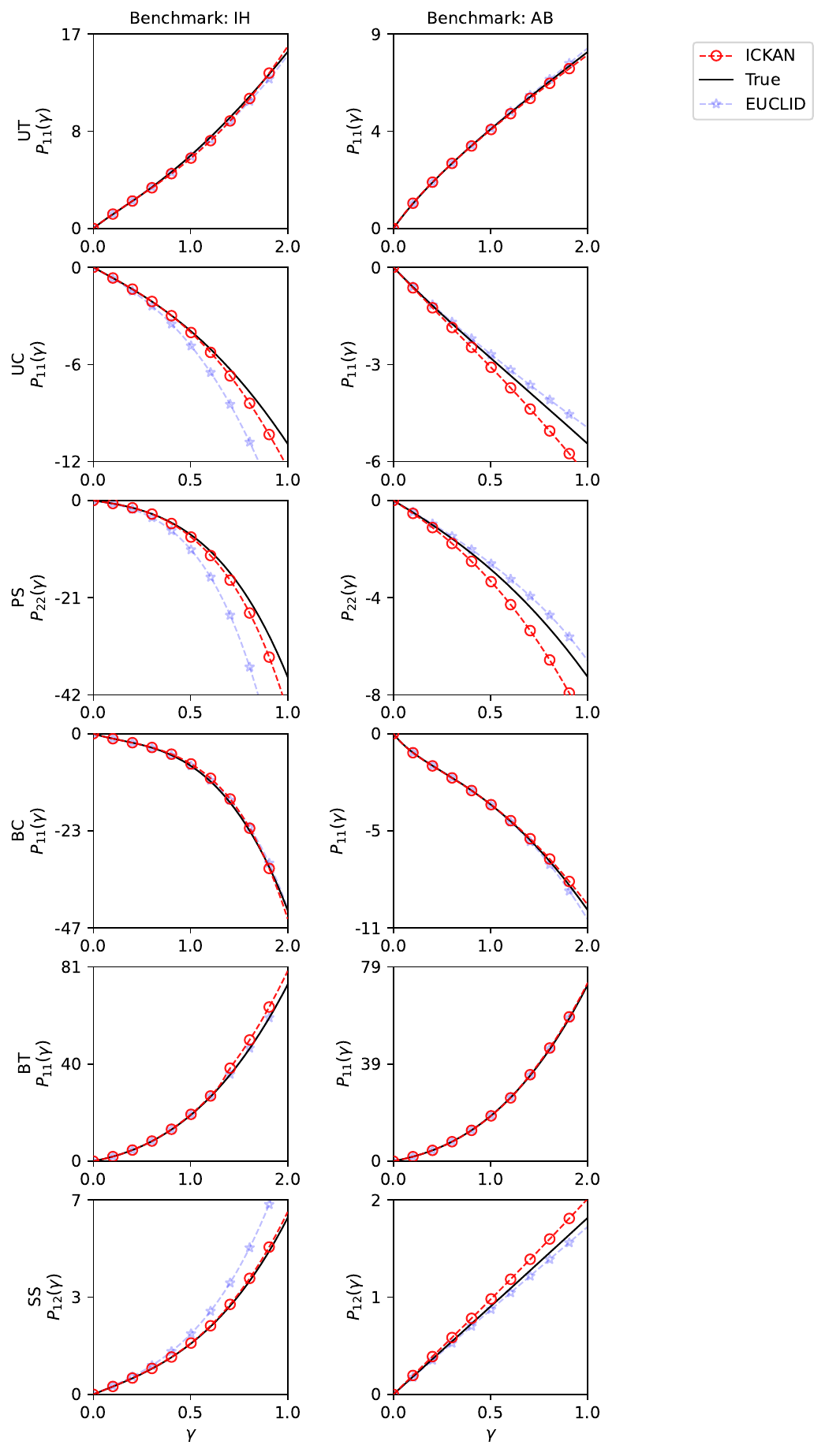}

\caption{\RV{Predicted first Piola-Kirchhoff stress $ \bfP(\bfF(\gamma)) $ components along the deformation paths specified in \eqref{eq:strain_paths} for the high noise case $(\sigma_u = 10^{-3})$. The best ICKAN-based constitutive model is shown for the IH \eqref{eq:IH} and AB \eqref{eq:AB} benchmarks, and the model obtained through EUCLID, alongside the response of the true (hidden) model for comparison.}}
\label{fig:EUCLID-IH_AB_noise=high_Pij}
\end{figure}
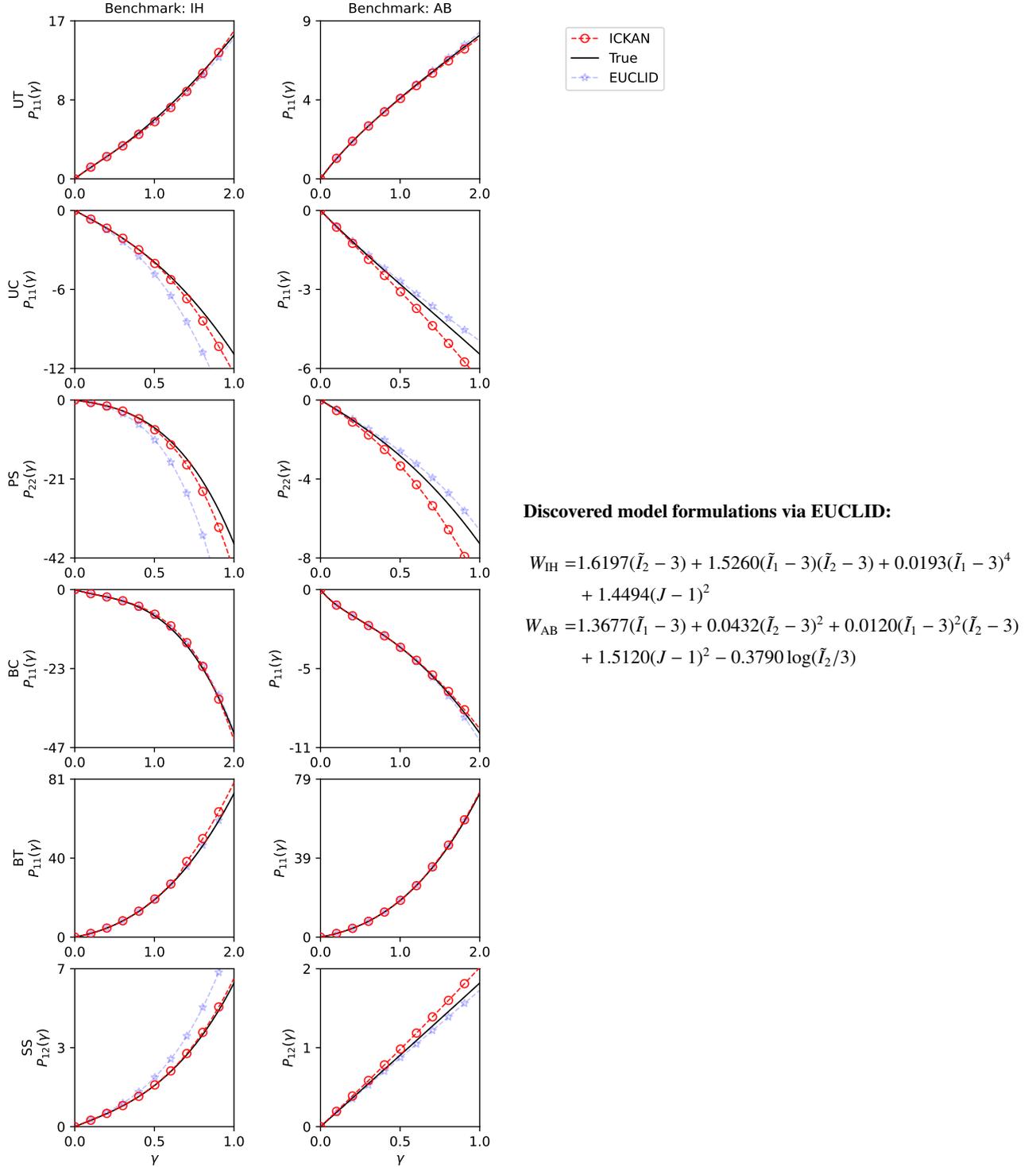

\RV{The results in \figurename~\ref{fig:EUCLID-IH_AB_noise=high_Pij} demonstrate that both the EUCLID framework \citep{flaschel_unsupervised_2021} and ICKAN perform well in capturing the underlying constitutive behavior for the IH \eqref{eq:IH} and AB \eqref{eq:AB} benchmarks under high noise conditions. EUCLID discovers physically meaningful and compact strain energy formulations by operating within a predefined functional basis, which promotes interpretability but can limit expressivity in more complex settings. In contrast, ICKANs begin with a highly expressive KAN-based architecture capable of fitting intricate stress–strain responses, and subsequently distills interpretable expressions through symbolic regression. For both benchmarks, the models obtained from EUCLID and ICKAN show strong agreement with the ground truth responses. These results highlight the complementary nature of the two approaches---EUCLID provides interpretable models by design, while ICKAN achieves accuracy-first learning with interpretability introduced in a post-processing stage.}

\section{\RV{Monotonic} input-convex symbolic regression}\label{sec:symbolic_regression_details}

To improve the interpretability of the learned ICKAN constitutive models, symbolic regression is applied to analytically approximate the trained univariate activation functions individually. The symbolic expressions of each activation are then assembled across the entire network to obtain a symbolic expression of the whole constitutive model. Here, we adapt the symbolic framework of \cite{liu2024kan} to enable \RV{monotonic} input-convex symbolic regression and, thereby, obtain symbolic polyconvex hyperelastic models.

Given a trained univariate function mapping from layer $r$ to layer  $r+1$, the activation function $\phi_{r,i,j}$ is approximated by searching for the best symbolic representation $\hat{\phi}_{r,i,j}$ from a candidate function library $\mathcal{W}$. In the subsequent discussion, we drop the subscripts on $\phi$ for the sake of brevity.

For a given candidate function $f\in\calW$ that is convex and non-decreasing, we construct an approximation ansatz of the form:
\be
\hat{\phi}(x) = c f(ax+b) + d, \qquad \text{with} \quad a\geq0 \quad \text{and} \quad c\geq0,
\ee
where $a,b,c,d$ are scalar fitting parameters. The convex and non-decreasing constraint on $f$ and along with $a\geq0$ and $c\geq0$ ensure that $\hat \phi$ is convex and non-decreasing. Consequently, this guarantees that the ICKAN's symbolic model is \RV{monotonic and} input-convex and the resulting hyperelastic strain energy density is polyconvex.

The parameters $a,b,c,d$ are obtained by numerically fitting $\hat\phi$ to the splines-based $\phi$ as
\be
a,b,c,d \leftarrow \arg\min_{a,b,c,d} \sum_{i=1}^{n}\|\hat\phi(x_i)-\phi(x_i)\|^2 \qquad \text{s.t.} \quad a\geq0 \quad \text{and} \quad  c\geq0.
\ee  
Here, the functions $\hat\phi$ and $\phi$ are evaluated at $n=100$  points for the purpose of fitting. The parameters $ a$ and $ b$ are determined via iterative grid search in the range $[0,10]$ and $[-10,10]$ respectively, while $ c$ and $ d$ are obtained through linear regression in each iteration.

In this study, we choose the function library as
\be
\RV{\calW = \{x, \exp(x), \log(1+\exp(x)), \log(1+\exp(x))^2, \log(1+\exp(x))^3, \log(1+\exp(x))^4\} \quad \text{with} \quad x\in\Rset,}
\ee
where all the candidate functions are convex and non-decreasing.

To select the optimal symbolic function $f$ from the library $\calW$, the following loss function and selection criterion is used:
\be  \label{eq:symbolic_loss}
f\leftarrow \arg\min_{f\in\calW} \  \lambda_{\text{sym}} \mathcal{L}_{\text{complexity}} + (1 - \lambda_{\text{sym}}) \mathcal{L}_{\text{fit}}\ ,  
\ee  
where $\lambda_\text{sym}\in[0,1]$ is a scalar hyperparameter. $ \mathcal{L}_{\text{fit}}$ quantifies the approximation error due to $f$ using the goodness-of-fit $ R^2$ score, defined as  
\be
\mathcal{L}_{\text{fit}} = \log_2(1 + 10^{-5} - R^2(\hat \phi, \phi)),
\ee  
while $ \mathcal{L}_{\text{complexity}}$ penalizes the complexity of $f$ (i.e., promotes parsimony) by assigning numerical weights to different expressions: 
\be
\mathcal{L}_\text{complexity}
=\begin{cases}
    1 \qquad \text{if} \quad f(x)=x,\\
    2 \qquad \text{if} \quad f(x)= \exp(x),\\
    \RV{2 \qquad \text{if} \quad f(x)=\log(1+\exp(x))^n, \quad \text{where } n\in{1,2,3,4}.}
\end{cases}
\ee
The parameter $ \lambda_\text{sym}$ is tunable (see Table~\ref{tab:parameters} for chosen value), which allows for a balance between simplicity and accuracy.

\end{document}